\documentclass[letterpaper, 10 pt, journal, twoside]{IEEEtran}
\IEEEoverridecommandlockouts     
\usepackage[english]{babel}
\usepackage{amsmath}
\usepackage{amssymb}
\usepackage{booktabs}
\usepackage{subcaption}
\usepackage[dvips]{graphicx}
\usepackage{multirow}
\usepackage{amsfonts}
\usepackage{enumerate}
\usepackage{tabularx}
\usepackage{algorithm,algorithmic}
\usepackage{bm}
\usepackage{enumerate}
\usepackage{array}
\usepackage{adjustbox}
\usepackage{xcolor}
\usepackage{siunitx}
\usepackage{booktabs}
\usepackage{mdframed}
\usepackage{adjustbox}
\usepackage{authblk}
\usepackage{color}
\usepackage{xurl}
\usepackage{pifont}
\usepackage{comment}
\usepackage{cite}
\makeatletter
\let\NAT@parse\undefined
\makeatother
\usepackage{hyperref}
\newcolumntype{s}{>{\hsize=.3\hsize}X}
\newcolumntype{x}{>{\hsize=.5\hsize}X}
\usepackage{tikz}
\usepackage{diagbox}

\newcommand{\cmark}{\textcolor{green}{\ding{51}}}\newcommand{\xmark}{\textcolor{red}{\ding{55}}}

\title{CoPeD-Advancing Multi-Robot Collaborative Perception: A Comprehensive Dataset in Real-World Environments}
\author{Yang Zhou$^{1}$, Long Quang$^{2}$, Carlos Nieto-Granda$^{2}$, and Giuseppe Loianno$^{1}$
\thanks{Manuscript received: March, 7, 2024; Accepted May, 21, 2024. This paper was recommended for publication by Editor M. Ani Hsieh upon evaluation of the Associate Editor and Reviewers' comments. This work was supported by the DARPA YFA Grant D22AP00156-00, the DEVCOM ARL grant SARA W911NF-24-2-0057 grant, and the NSF CPS Grant CNS-2121391.}
\thanks{$^1$The authors are with the New York University, Tandon School of Engineering, 11201 Brooklyn, NY, USA. {\tt\footnotesize email: \{yangzhou, loiannog\}@nyu.edu}.}
\thanks{$^2$The authors are with the  U.S. Army Combat
Capabilities Development Command, Army Research Laboratory, Adelphi, MD 20783, USA. {\tt\footnotesize email: \{long.p.quang.civ, carlos.p.nieto2.civ\}@army.mil}.}
\thanks{The authors would like to thank Julie Foresta, Nishanth Bobbili, Roshan Balu T. M. B., Luca Morando, Jeffery Mao, and Guanrui Li for their help during the dataset collection process.}
\thanks{Digital Object Identifier (DOI): see top of this page.}
}
\begin{document}
\IEEEpubid{\begin{minipage}[t]{\textwidth}\ \\[10pt]        {\copyright 2024 IEEE. Personal use is permitted, but republication/redistribution requires IEEE permission. See https://www.ieee.org/publications/rights/index.html for more information.}\end{minipage}} 
\maketitle

\begin{abstract}
In the past decade, although single-robot perception has made significant advancements, the exploration of multi-robot collaborative perception remains largely unexplored. This involves fusing compressed, intermittent, limited, heterogeneous, and asynchronous environmental information across multiple robots to enhance overall perception, despite challenges like sensor noise, occlusions, and sensor failures. One major hurdle has been the lack of real-world datasets.
This paper presents a pioneering and comprehensive real-world multi-robot collaborative perception dataset to boost research in this area. Our dataset leverages the untapped potential of air-ground robot collaboration featuring distinct spatial viewpoints, complementary robot mobilities, coverage ranges, and sensor modalities. It features raw sensor inputs, pose estimation, and optional high-level perception annotation, thus accommodating diverse research interests. Compared to existing datasets predominantly designed for Simultaneous Localization and Mapping (SLAM), our setup ensures a diverse range and adequate overlap of sensor views to facilitate the study of multi-robot collaborative perception algorithms. We demonstrate the value of this dataset qualitatively through multiple collaborative perception tasks. We believe this work will unlock the potential research of high-level scene understanding through multi-modal collaborative perception in multi-robot settings.
\end{abstract}
\begin{IEEEkeywords}
Data Sets for Robotic Vision, Deep Learning for Visual Perception, Multi-Robot Systems
\vspace{-10pt}
\end{IEEEkeywords}
\section*{Supplementary Material}
\noindent \textbf{Video}: \url{https://youtu.be/3xGmo0mjVzg}\\
\noindent \textbf{Dataset}: \url{https://github.com/arplaboratory/CoPeD}

\begin{figure}[t]
    \centering
\begin{subfigure}[b]{0.5\linewidth}
\includegraphics[height=0.6\textwidth, width=\textwidth]{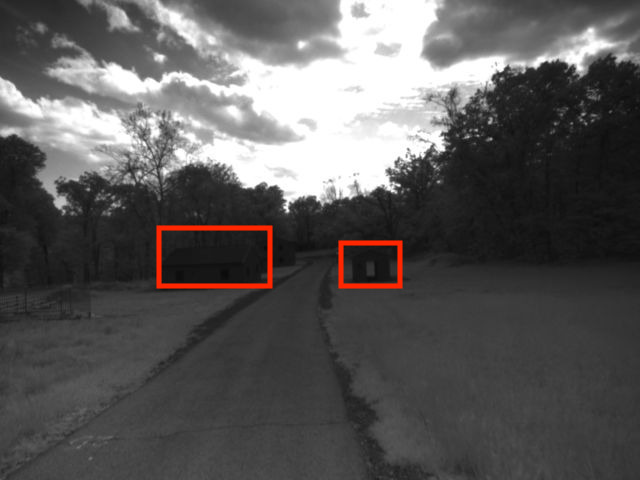}
    \end{subfigure}\begin{subfigure}[b]{0.5\linewidth}
\includegraphics[height=0.6\textwidth, width=\textwidth]{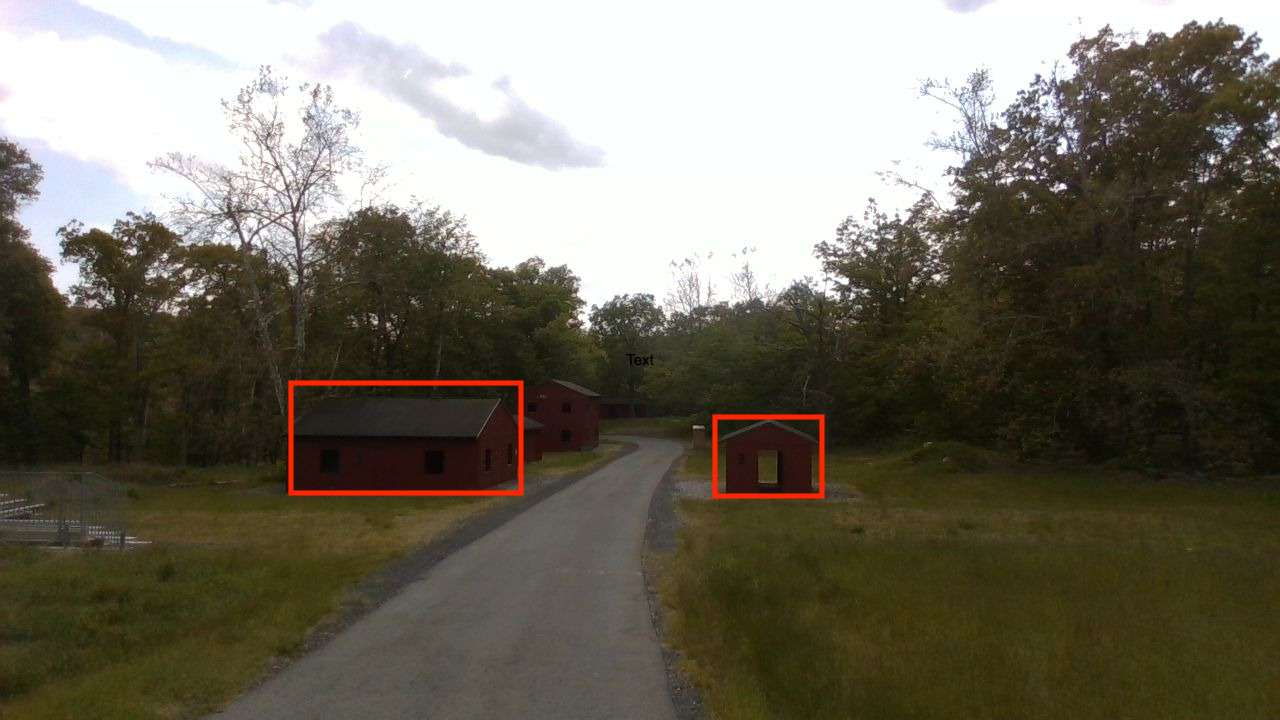}
    \end{subfigure}
    \begin{subfigure}[b]{\linewidth}
\includegraphics[height=0.6\textwidth, width=\textwidth]{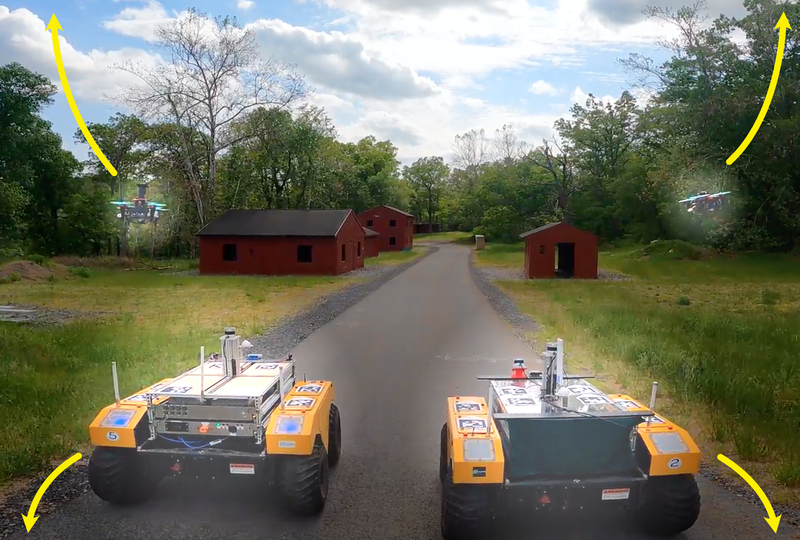}
    \end{subfigure}
    \begin{subfigure}[b]{0.5\linewidth}
\includegraphics[height=0.6\textwidth, width=\textwidth]{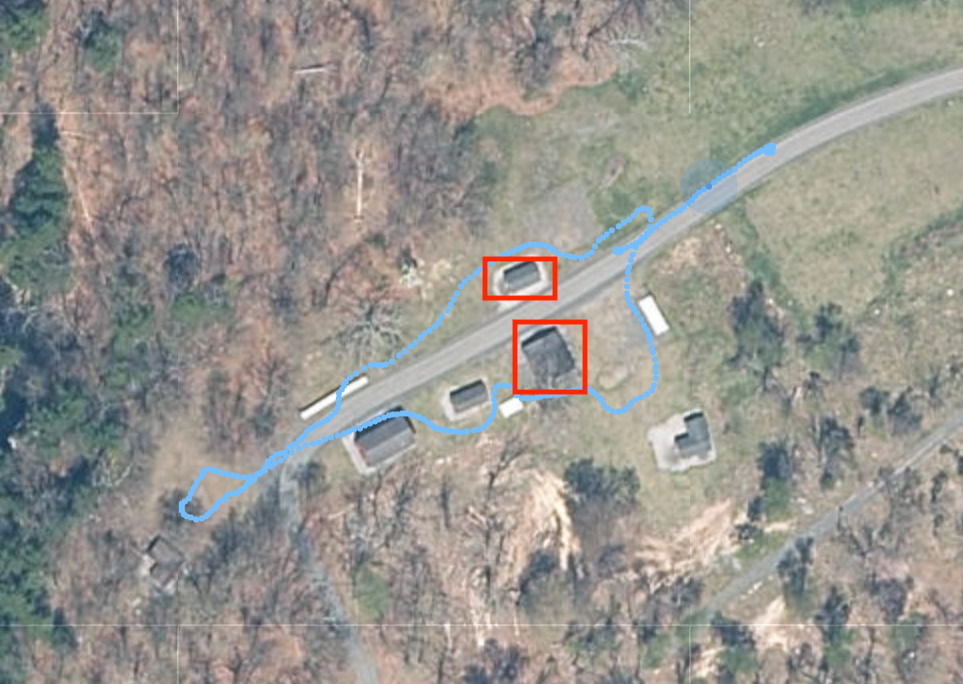}
    \end{subfigure}\begin{subfigure}[b]{0.5\linewidth}
\includegraphics[height=0.6\textwidth, width=\textwidth]{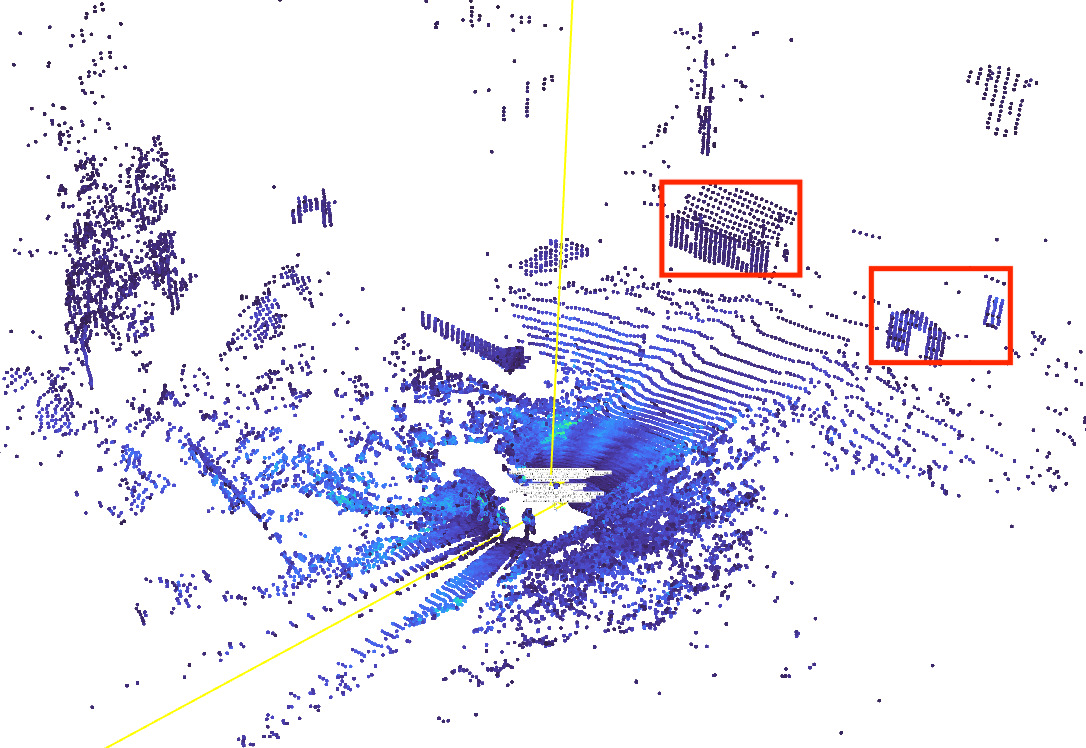}
    \end{subfigure}
    \caption{Multi-robot collaborative data collection. The top row shows the infrared and RGB images from the aerial robots, whereas the bottom row shows the GPS and LiDAR data from the ground robots. The sensor data present several overlapping spatial areas. Red boxes show sample objects identified in multiple heterogeneous sensors' data streams.}
    \label{fig:teaser}
    \vspace{-10pt}
\end{figure}
\begin{table*}[t]
\centering
\caption{Taxonomy of existing datasets.}
\begin{tabularx}{\linewidth}{c X c c c c }
\toprule
\textbf{Task} & \textbf{Approaches} & \textbf{Platform Type} & \textbf{Real-World} & \textbf{Modalities} & \textbf{Scene} \\
\midrule
\multirow{5}{*}{Multi-Robot SLAM}
 & Lamp 2.0~\cite{chang2022lamp} & Multi-Ground Robot & \cmark & Camera, LiDAR & Outdoor \\
\cmidrule{2-6}
& Subt-Cerberus~\cite{tranzatto2022team} & Multi-Ground Robot & \cmark & Camera, LiDAR, IMU & Outdoor \\
\cmidrule{2-6}
& Graco~\cite{zhu2023graco} & Aerial/Ground Robot & \cmark & Camera, LiDAR, IMU, GPS & Outdoor \\
\cmidrule{2-6}
& S3e~\cite{feng2022s3e} & Aerial/Ground Robot & \cmark & Camera, LiDAR, IMU, GPS & Indoor/Outdoor \\
\cmidrule{2-6}
& Tian et al.~\cite{tian2023resilient} & Multi-Ground Robot & \cmark & Camera, LiDAR, IMU, GPS & Outdoor \\
\midrule
\multirow{7}{*}{Collaborative Perception} & Opv2v~\cite{xu2022opv2v} & Vehicle-to-Vehicle (V2V) & \xmark & Camera, LiDAR, IMU, GPS & Outdoor\\
\cmidrule{2-6}
& Arnold et al.~\cite{arnold2020cooperative} & Vehicle-to-Vehicle (V2V) & \xmark & Camera, Depth Camera & Outdoor \\
\cmidrule{2-6}
& Dair-V2X~\cite{yu2022dair} & Vehicle-to-Everything (V2X) & \cmark & Camera, LiDAR, IMU, GPS & Outdoor \\
\cmidrule{2-6}
& V2X-Sim~\cite{li2022v2x} & Vehicle-to-Vehicle (V2V) & \xmark & Camera, LiDAR, IMU, GPS & Outdoor \\
\cmidrule{2-6}
& When2comm~\cite{liu2020when2com} & Multi-Aerial Robot & \xmark & Camera & Outdoor \\
\cmidrule{2-6}
& Where2comm~\cite{hu2022where2comm} & Multi-Aerial Robot & \xmark & Camera & Outdoor \\
\cmidrule{2-6}
& CoPeD (Ours) & Multi-Aerial/Ground Robot & \cmark & Camera, LiDAR, IMU, GPS & Indoor/Outdoor \\
\bottomrule
\end{tabularx}
\label{tab:related}
\vspace{-10pt}
\end{table*}

\section{Introduction}~\label{sec:introduction}
\vspace{-10pt}
\IEEEPARstart{R}{obotic} perception has advanced significantly over the past decade, largely due to progress in machine learning and the widespread adoption of increasingly powerful computing and sensor systems. However, most of these developments have been centered on single-robot systems, leaving the area of multi-robot collaborative perception relatively unexplored. This form of perception refers to a network of robots, each with its unique sensing characteristics, capabilities, and spatial viewpoints being able to collect and fuse environmental sensing data in an efficient, distributed, and meaningful way to obtain accurate context-appropriate information, leading to potential advantages in terms robustness, accuracy, and redundancy compared to a single robot solution. Pushing the frontier of multi-robot collaborative perception could harness these unique advantages to create a more comprehensive, unified, and resilient perception system. The creation of multi-robot perception datasets can support and facilitate the development of collaborative perception models and algorithms. In this paper, we present the first heterogeneous multi-robot perception dataset collected in challenging indoor and outdoor settings using ground and aerial robots. 

Notably, the multi-robot air-ground systems have the potential to provide a distinct opportunity for multi-robot perception. Specifically, ground and small-scale aerial robots combine different mobility and sensing characteristics (e.g., types of sensors as well as different spatial viewpoints and sensing range characteristics). For instance, ground robots can carry heavy payloads such as 3D LiDAR, whereas small-scale aerial robots due to their Size, Weight, and Power (SWaP) constraints generally can only rely on small sensors such as cameras and Inertial Measurements Units (IMUs).

Existing multi-robot datasets are mainly geared towards Simultaneous Localization and Mapping (SLAM) applications~\cite{chang2022lamp, tranzatto2022team, zhu2023graco, feng2022s3e, tian2023resilient}, focusing primarily on spatial coverage rather than collaborative perception. The limited overlap of sensor views in these datasets hinders the study of perception results that rely on perception compensation mechanisms between robots. Furthermore, most of the existing multi-agent perception datasets are based on photorealistic simulators, neglecting the real-world sensor noise and characteristics inherent in real-world multi-robot settings. 
Although there are datasets targeting autonomous driving systems through vehicle-to-vehicle (V2V) communication~\cite{xu2022opv2v,arnold2020cooperative,yu2022dair,li2022v2x} these fall short due to the mismatch between simulated and real-world data or the absence of multiple vehicles with heterogeneous mobility and modalities. Furthermore, the algorithms developed from these datasets are designed for Bird's Eye View (BEV) since the scene is constrained on the flat surface and agent mobility is limited to ground vehicle. Due to the limited scope and biases of existing datasets, there is no existing methods that utilize multiple modalities, multiple mobilities and goes beyond BEV paradigm. Creating a comprehensive dataset tailored to address multi-robot collaborative perception problems, a crucial yet unexplored area in the community, is indispensable to making significant progress in this area and push the frontier to develope new methods.

The contributions of this paper can be summarized as follows. First, we present the first multi-robot dataset tailored to collaborative perception tasks featuring both indoor and outdoor sequences, multiple sensor spatial viewpoints, robot types, multi-rate and multi-modal data as well as coverage ranges. It reflects the real-world sensor noise and disturbances encountered when deploying a heterogeneous robot team in the wild.
The aerial robots equipped with stereo cameras, forward and downward RGB cameras, depth sensors, Inertial Measurement Units (IMUs), and GPS. Ground robots carry a stereo camera, forward RGB camera, depth sensor, 3D LiDAR, IMU, and GPS. 
Second, to cater to diverse research needs, in addition to raw sensor data streams, we provide pose estimation and optional high-level perception annotation. Specifically, pose estimation is obtained by fusing GPS data with existing SLAM frameworks. For high-level perception annotation, we employ foundation models to obtain zero-shot automatic annotation. Finally, we validate the usefulness of the dataset in several collaborative perception tasks. Our dataset is a first-of-its-kind contribution to the robotics community going beyond the BEV autonomous driving scenario. The incorporation of multiple sensor modalities and robots enables to pave the way for studying multi-modal collaborative perception using real-world data.

The paper is organized as follows. In Section~\ref{sec:relatedworks}, we discuss existing datasets. In Section~\ref{sec:methodology}, we present the dataset characteristics, and in Section~\ref{sec:calibration} the technical details related to sensors and calibration. Section~\ref{sec:data} introduces the pose and perception annotations procedures, Section~\ref{sec:attributes} discuss the dataset attributes, whereas Section~\ref{sec:usecases} presents use cases and applications. Finally, Section~\ref{sec:conclusion} concludes the work and presents several future research directions.

\begin{figure*}[t]
    \centering
    \begin{subfigure}[t]{0.325\linewidth}
         \centering
         \includegraphics[width=\textwidth, height=4cm]{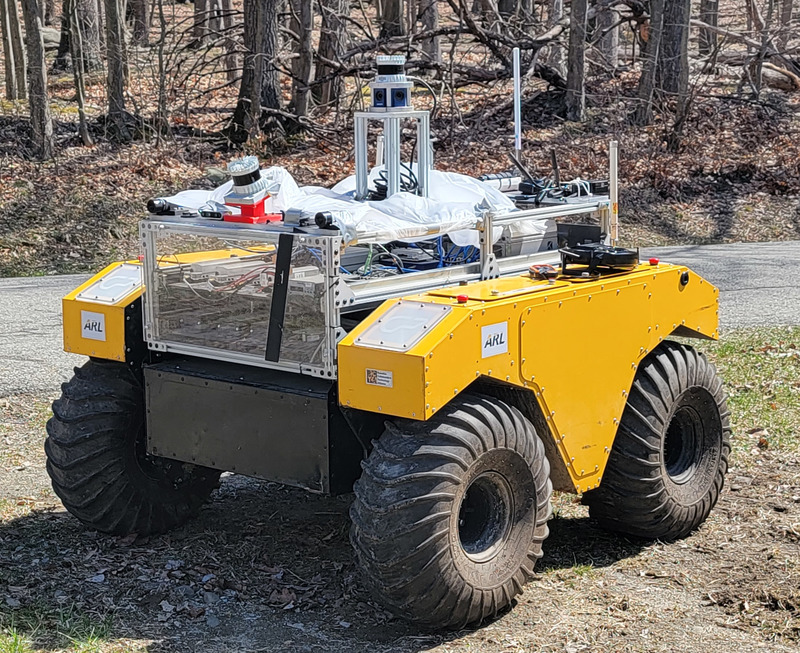}
         \caption{ARL Warthog Ground Robot}
\end{subfigure}
     \begin{subfigure}[t]{0.325\linewidth}
         \centering
         \includegraphics[width=\textwidth, height=4cm]{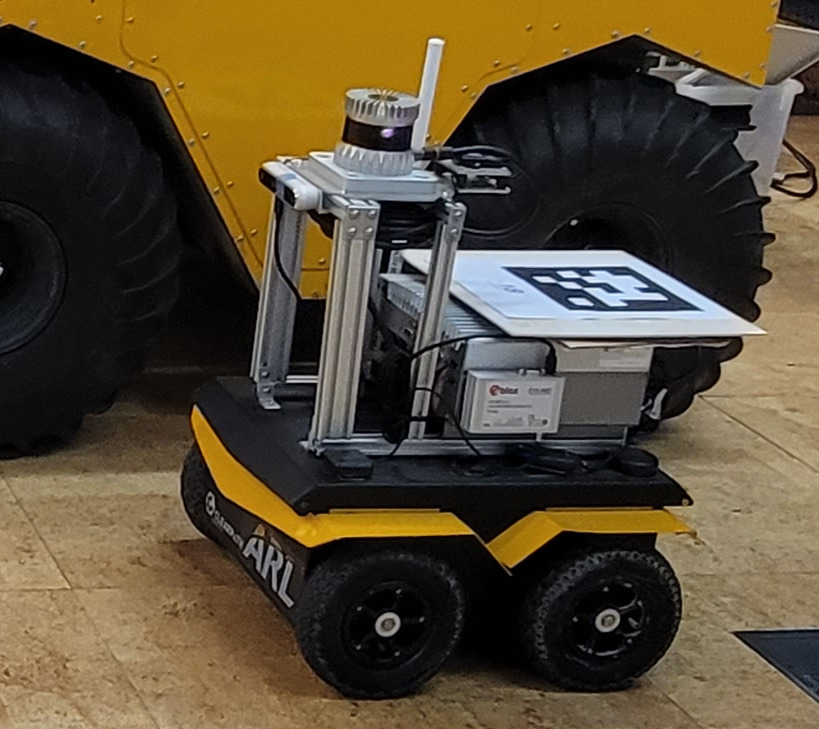}
         \caption{ARL Jackal Ground Robot}
\end{subfigure}
\begin{subfigure}[t]{0.325\linewidth}
         \centering
         \includegraphics[width=\textwidth, height=4cm]{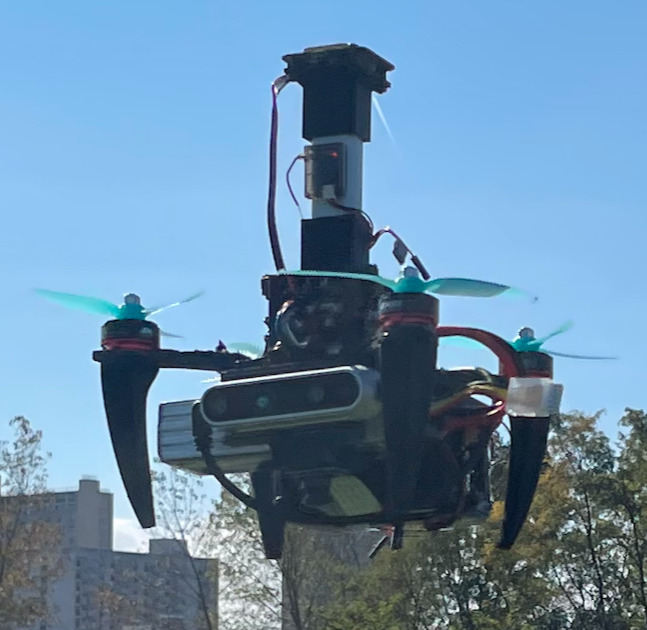} \caption{NYU ARPL Race Aerial Robot}
\end{subfigure}
     \caption{Layout of the ground robots and aerial robots.}
     \label{fig:robotlayouts}
     \vspace{-10pt}
\end{figure*}

\section{Related Works}~\label{sec:relatedworks}
This section reviews existing datasets and highlights the differences between our approach and existing solutions as illustrated in Table~\ref{tab:related}. The field of multi-robot collaborative perception is relatively new~\cite{zhou2022multi,li2023multi}, with few datasets available for research purposes within the community. Existing datasets are primarily designed for SLAM applications~\cite{chang2022lamp, tranzatto2022team, zhu2023graco, feng2022s3e, tian2023resilient}, while our dataset is specifically tailored to facilitate research into multi-robot collaborative perception. 

Historically, the multi-robot collaborative perception was primarily confined to geometric mapping in robotics. However, significant advancements have been made in this area by the research community~\cite{lajoie2021towards}, resulting in the release of several multi-robot SLAM datasets~\cite{chang2022lamp, tranzatto2022team,zhu2023graco,feng2022s3e,tian2023resilient}. These datasets contain outdoor sequences, with the exception of~\cite{feng2022s3e}, which includes both indoor and outdoor sequences. Datasets such as~\cite{chang2022lamp,feng2022s3e,tian2023resilient} deploy multiple ground robots for data collection, and~\cite{tranzatto2022team,zhu2023graco} extend this to air-ground robot teams, but only in outdoor urban settings. Despite these advancements, these datasets are primarily designed for SLAM applications, focusing on geometric mapping rather than high-level collaborative perception tasks, including monocular depth estimation~\cite{zhou2022multi}, semantic segmentation~\cite{liu2020when2com,liu2020who2com, zhou2022multi,fan2021few}, correspondence matching~\cite{gao2023deep}, object detection~\cite{hu2022where2comm,li2023multi,li2021learning,xu2022bridging,lei2022latency, cui2022coopernaut}. The limited overlap of sensor views and types of environments in these datasets hinders the study of perception models and algorithms that exploit the benefits of using multiple robots.

There are a few collaborative perception datasets target autonomous driving systems through V2V communication~\cite{xu2022opv2v,arnold2020cooperative,yu2022dair,li2022v2x}. Specifically, most of these datasets, due to the difficulties associated with the collection and annotation of real-world data, are based on simulation with the exception of~\cite{yu2022dair} which provides the first real-world V2X dataset. The dataset from~\cite{arnold2020cooperative} employs LiDAR as the single sensing modality, while~\cite{xu2022opv2v,yu2022dair,li2022v2x} adopt multiple modalities including cameras and 3D LiDARs. However, these datasets are intended for autonomous driving applications, missing the aerial complementary viewpoints and perspectives, and focusing on 3D object detection tasks. In addition, most of these datasets are based on simulation due to the difficulties associated with real-world data collection and annotation. In addition to autonomous driving-focused datasets,~\cite{liu2020when2com} and~\cite{hu2022where2comm} provide simulation-based photorealistic datasets from multiple aerial robot perspectives. Therefore, it is important to acknowledge that this data alone cannot completely replicate the nuanced aspects of robot data available in the real-world. Finally, existing datasets do not fully offer the benefits of data streams from multiple distinct spatial complementary viewpoints and perspectives as in the proposed air-ground case.

Datasets need to be more comprehensive to be used for collaborative perception tasks. Very few are designed for heterogeneous multi-robot applications, and the sensor modality setups are unsuitable for multi-robot collaborative perception. Compared to existing datasets as summarized in Table~\ref{tab:related}, the proposed dataset is the first real-world heterogeneous air-ground collaborative perception dataset, designed explicitly for multi-modal collaborative robot perception research.
\begin{table*}
    \centering
    \caption{Ground Robot Sensor Specifications.}
    \begin{tabular}{>{\centering\arraybackslash}p{0.1\linewidth} >{\centering\arraybackslash}p{0.18\linewidth} >{\centering\arraybackslash}p{0.29\linewidth} >{\centering\arraybackslash}p{0.1\linewidth} >{\centering\arraybackslash}p{0.1\linewidth} >{\centering\arraybackslash}p{0.1\linewidth} }
        \toprule
        Equipment & Model Type & Characteristics & Resolution & FoV & Sensor Rate\\
        \midrule
        LiDAR & Ouster OS1-64 & $200~\si{m}$ range& $512\times 64$ & $45 ^{\circ}$ vertical & $10$ Hz \\
        \midrule
        LiDAR & Ouster OS1-128 & $200~\si{m}$ range & $512\times 128$ & $45 ^{\circ}$ vertical & $10$ Hz \\
        \midrule
        LiDAR & Ouster OS0-64 & $100~\si{m}$ range & $512\times 64$ & $90 ^{\circ}$ vertical & $10$ Hz \\
        \midrule
        \multirow{2}{*}{IMU} & LORD Microstrain 3DM-GX5-25 & \multirow{2}{*}{$\pm 8$ g}& \multirow{2}{*}{$300$ dps}  & - & \multirow{2}{*}{$1000$ Hz} \\ \midrule
        GNSS & u-Blox EVK-M8T & $2~\si{m}$, $0.3 ^{\circ}$ accuracy & - & - & $2$ Hz\\ \midrule
        & \multirow{3}{*}{Intel Realsense D435i} & RGB: Rolling Shutter & $1920\times1080$ & $69^{\circ} \times 42^{\circ}$ & $30$ Hz  \\ RGBD & & Stereo IR: Global Shutter, Baseline: $50$ mm & $640\times480$ & $87^{\circ} \times 58^{\circ}$ & $30$ Hz \\ 
        (Wilbur) & & IMU: BMI 055 & - & - & $200$ Hz \\
        \midrule
        & \multirow{3}{*}{Intel Realsense D455} & RGB: Global Shutter & $1280\times 800$ & $90^{\circ} \times 65^{\circ}$ & $30$ Hz  \\ 
        RGBD & & Stereo IR: RS, Baseline: $95~\si{mm}$& $1280\times720$ & $87^{\circ} \times 58^{\circ}$ & $30$ Hz \\ 
        (Wanda) & & IMU: BMI 055& - & - & $200$ Hz  \\
        \midrule
        \multirow{2}{*}{Stereo Camera} & FLIR Blackfly S (BFS-PGE-16S2C-CS) &  \multirow{2}{*}{RGB: Global Shutter, Baseline: $60$ cm} & \multirow{2}{*}{$1440\times 1080$}  & \multirow{2}{*}{-} & \multirow{2}{*}{$78$ Hz}\\ \midrule
        Sync & Masterclock GMR1000 & $\pm 3$ second / year & - & - & - \\
        \midrule
        \multirow{2}{*}{Main Computer} & \multirow{2}{*}{Nuvo-7166GC} &Intel i7-8700 CPU @ 3.20GHz w. 32GB RAM, NVIDIA T4 GPU w. 16GB RAM & - & - & -\\
        \midrule
        \multirow{2}{*}{Vision Computer} & \multirow{2}{*}{Nuvo-7166GC} & Intel i7-8700 CPU @ 3.20GHz w. 32GB RAM, NVIDIA T4 GPU w. 16GB RAM & - & - & -\\
        \bottomrule 
    \end{tabular}
    \label{tab:groundrobotsensors}
\end{table*}
\begin{table*}
    \centering
    \caption{Aerial Robot Sensor Specifications.}
    \begin{tabular}{>{\centering\arraybackslash}p{0.08\linewidth} >{\centering\arraybackslash}p{0.18\linewidth} >{\centering\arraybackslash}p{0.29\linewidth} >{\centering\arraybackslash}p{0.1\linewidth} >{\centering\arraybackslash}p{0.1\linewidth} >{\centering\arraybackslash}p{0.1\linewidth} }
        \toprule 
        Equipment & Model Type & Characteristics & Resolution & FoV & Sensor Rate \\
        \midrule
        \multirow{3}{*}{GNSS} & \multirow{3}{*}{mRO Location One} & ublox NEO-M9N & - & - & $8$ Hz \\ & & Magnetic Compass RM3100& - & - & $8$ Hz \\ & &  Barometer DPS310 & - & - & $8$ Hz \\ \midrule\newcommand\boldblue[1]{\textcolor{blue}{\textbf{#1}}}
        RGB & IMX219 & RGB: Global Shutter & $640\times480$ & $160^{\circ} \times 160^{\circ}$ &$30$ Hz \\\midrule
        \multirow{3}{*}{RGBD} & \multirow{3}{*}{Intel Realsense D435i} & RGB: Rolling Shutter & $1920\times1080$ & $69^{\circ} \times 42^{\circ}$ & $30$ Hz  \\ 
        & & Stereo IR: Global Shutter, Baseline: $50~\si{mm}$ & $640\times480$ & $87^{\circ} \times 58^{\circ}$ & $30$ Hz \\ 
        & & IMU: BMI 055 & - & - & $200$ Hz \\
        \midrule
        \multirow{3}{*}{Computer} & \multirow{3}{*}{NVIDIA Jetson Xavier NX} & 6-core NVIDIA Carmel ARM v8.2 64-bit CPU, 384-core NVIDIA Volta GPU with 48 Tensor Cores, 8 GB RAM & \multirow{3}{*}{-} & \multirow{3}{*}{-} & \multirow{3}{*}{-}\\
        \bottomrule 
    \end{tabular}
    \label{tab:aerialrobotsensors}
    \vspace{-10pt}
\end{table*}

\section{Dataset Description}~\label{sec:methodology}
\vspace{-20pt}
\subsection{Robots and Equipment}
Our dataset utilizes three ground robots and two aerial robots. The robot layouts are illustrated in Fig.~\ref{fig:robotlayouts}. The Clearpath Robotics Warthog platform is equipped with one Ouster OS1-64 3D LiDAR, a LORD Microstrain 3DM-GX5-25 IMU, an u-Blox EVK-M8T GNSS module, an Intel Realsense D435i/D455 stereo camera, two high-resolution FLIR Blackfly S cameras and a hardware time synchronization Masterclock GMR1000. The Clearpath Robotics Jackal platform is equipped with an Ouster OS1-64 3D LiDAR, a LORD Microstrain 3DM-GX5-
25 IMU, an u-Blox EVK-M8T GNSS module, and an Intel
Realsense D435i/D455 stereo camera.
Each of the aerial robots, developed at the Agile Robotics and Perception Lab (ARPL) \footnote{\url{https://wp.nyu.edu/arpl/}} at the New York University, carries a forward Intel Realsense D435i stereo camera, a downward IMX219 color camera, a top-mounted PX4 autopilot module and an mRO Location One GPS module. The D435i stereo camera includes an RGB camera module, a stereo infrared camera module, and an IMU module. 
The complete list of sensor specifications is shown in Tables~\ref{tab:groundrobotsensors} and~\ref{tab:aerialrobotsensors}.  All robots' autonomy software stack builds upon ROS 1~\cite{quigley2009ros}\footnote{\url{http://wiki.ros.org/}}, and all robots are connected under one 5G Wi-Fi subnetwork with time synchronization to a laptop's system clock using Network Time Protocol (NTP). 

We mount AprilTag~\cite{olson2011apriltag} on ground robots illustrated in Fig.~\ref{fig:robotlayouts} to provide relative localization visual markers between the aerial robots and the ground robots. Compared to other visual fiducial markers, AprilTags guarantee increased robustness and accuracy from the perspectives of long-range detection and localization accuracy under real-time computation constraints. The aerial robots obtain the relative pose of the ground robots by AprilTag detection and Perspective-n-Point (PnP)~\cite{collins2014infinitesimal} using downward cameras. We use $8$ AprilTags as a tag bundle in order to facilitate robust AprilTag detection in the presence of occlusion, illumination changes, motion blur, and FoV constraints. The tag bundle is calibrated and detected using the method in \cite{malyuta2018guidance}.

Our heterogeneous robot team setup ensures that a diverse array of sensor modalities and robot mobilities are captured, thus enriching the dataset and providing more robust material for multi-modal heterogeneous robotic perception studies.
\begin{figure}[t]
    \centering
    \rotatebox{90}{\parbox{0.24\linewidth}{\centering \scriptsize PointCloud}}
    \begin{subfigure}[b]{0.47\linewidth}
         \centering
         \includegraphics[width=\textwidth, height=0.6\textwidth]{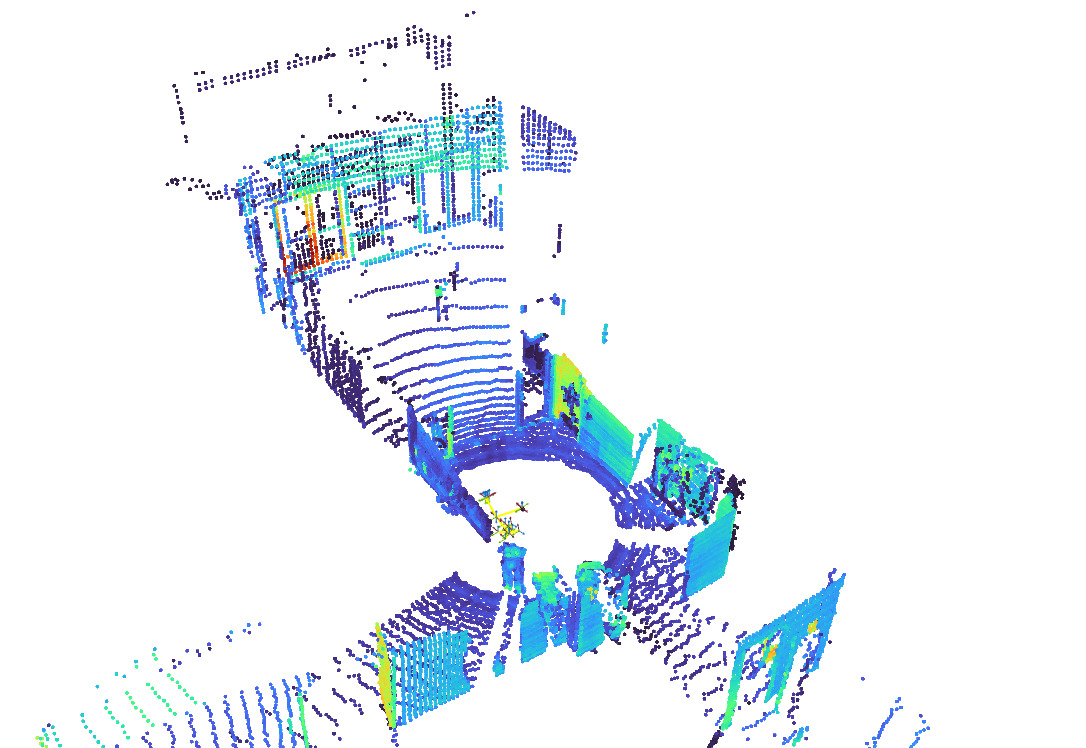}
\end{subfigure}
     \begin{subfigure}[b]{0.47\linewidth}
         \centering
         \includegraphics[width=\textwidth, height=0.6\textwidth]{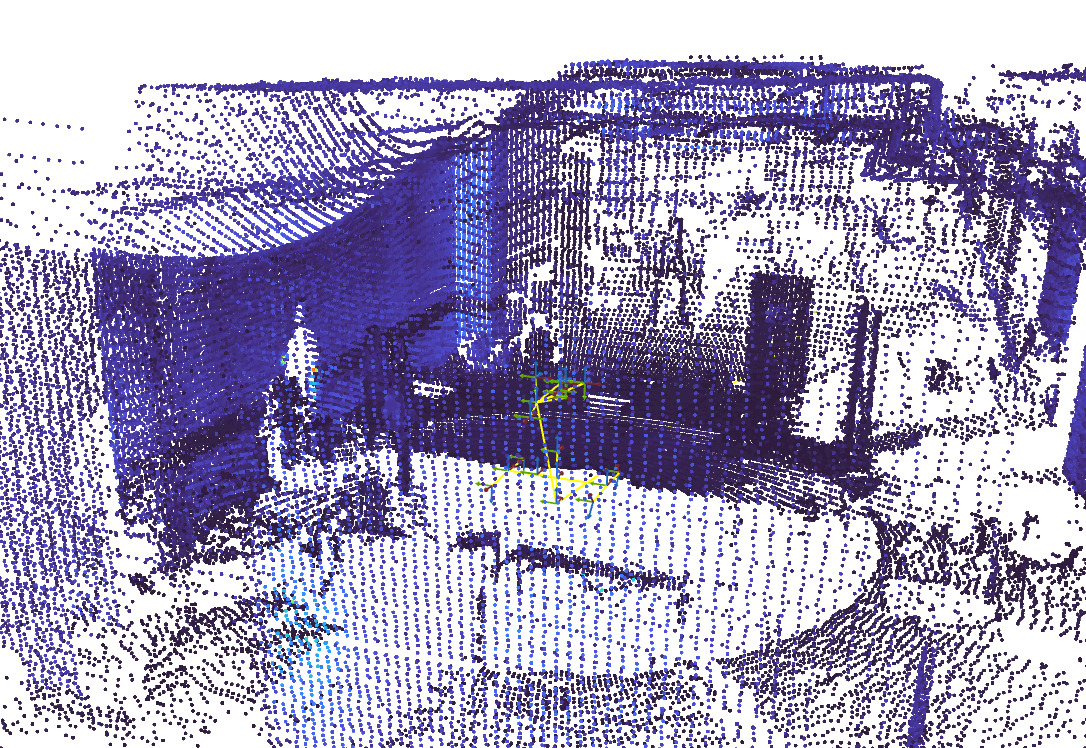}
\end{subfigure}
\rotatebox{90}{\parbox{0.24\linewidth}{\centering \scriptsize RGB}}
     \begin{subfigure}[b]{0.47\linewidth}
         \centering
         \includegraphics[width=\textwidth]{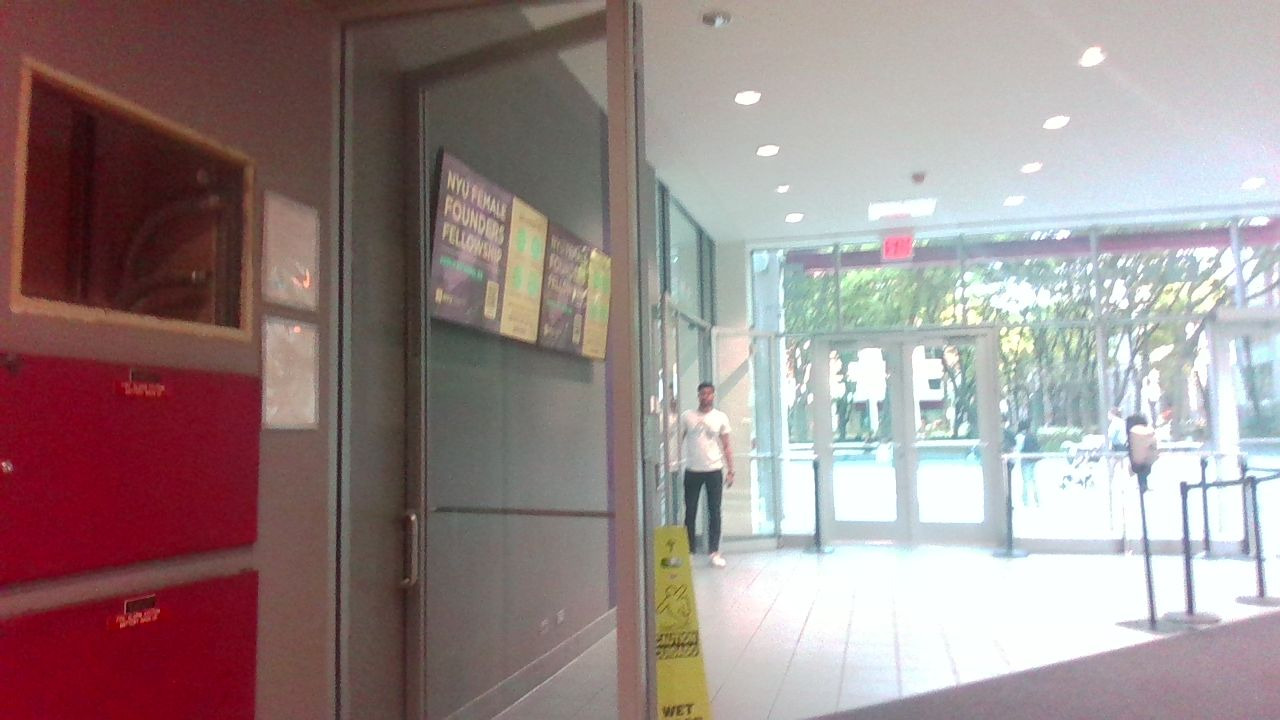}
\end{subfigure}
     \begin{subfigure}[b]{0.47\linewidth}
        \centering
        \includegraphics[width=\textwidth]{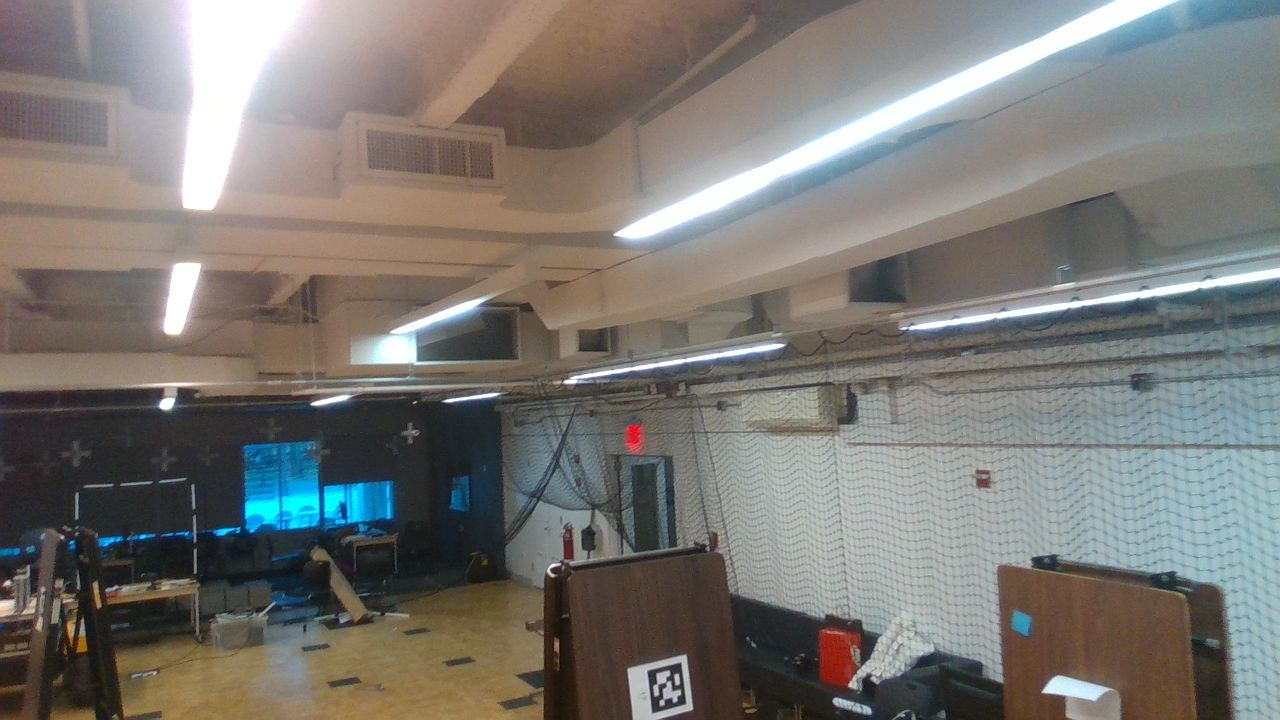}
\end{subfigure}
    \caption{Indoor-NYUARPL environment. Pointcloud data (top row) is captured by the ground robots, and the color images (bottom row) are captured by the aerial robots. Each column shows one subgroup of one aerial robot and one ground robot exploring different parts of the environment.}
    \label{fig:samples-GAMELAB}
\end{figure}

\subsection{Indoor Sequences}
For indoor sequences, we use one Jackal ground robot, one Warthog ground robot, and two aerial robots. We present three sequences with one Warthog ground robot and one aerial robot, and one sequence with two ground robots and two aerial robots as shown in Fig.~\ref{fig:samples-GAMELAB}. 

The indoor sequences named 'Indoor-NYUARPL' are captured in a large indoor environment of $38~\si{m} \times 60 ~\si{m}$ with a rich composition of objects including tables, chairs, and doors. 
We split the robot team into two subteams, each of them consisting of one aerial robot and one ground robot. The two subteams explore different parts of the environment. The aerial robots fly at a height of $2.0~\si{m}$ above the ground robots, whereas the ground robots move at a speed of $0.5~\si{m/s}$. The first team explores a lab room and the second one explores a hallway characterized by natural lighting. 

\begin{figure}[t]
    \centering
    \rotatebox{90}{\parbox{0.24\linewidth}{\centering \scriptsize PointCloud}}
    \begin{subfigure}[b]{0.47\linewidth}
         \centering
         \includegraphics[width=\textwidth, height=0.6\textwidth]{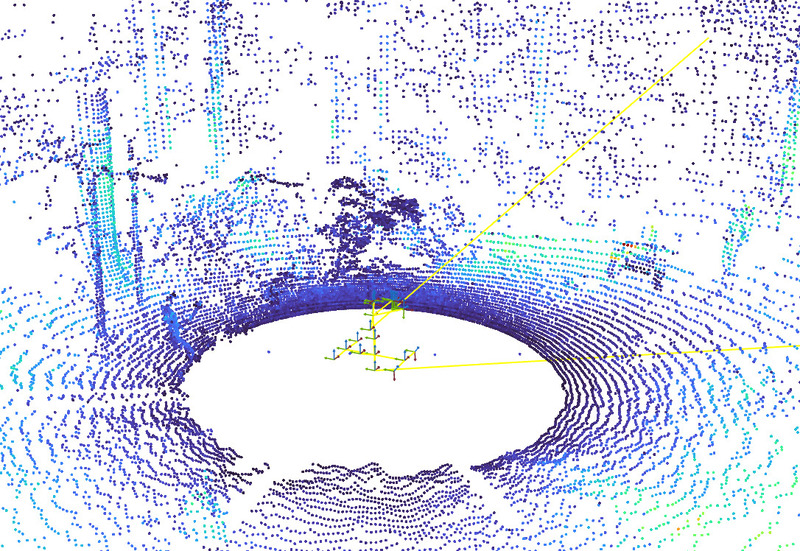}
     \end{subfigure}
     \begin{subfigure}[b]{0.47\linewidth}
         \centering
         \includegraphics[width=\textwidth, height=0.6\textwidth]{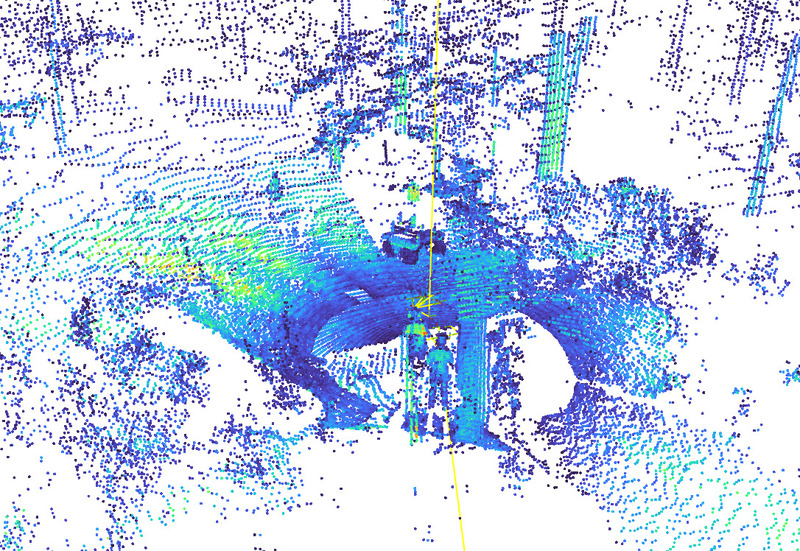}
     \end{subfigure}
\rotatebox{90}{\parbox{0.24\linewidth}{\centering \scriptsize RGB}}
     \begin{subfigure}[b]{0.47\linewidth}
         \centering
         \includegraphics[width=\textwidth]{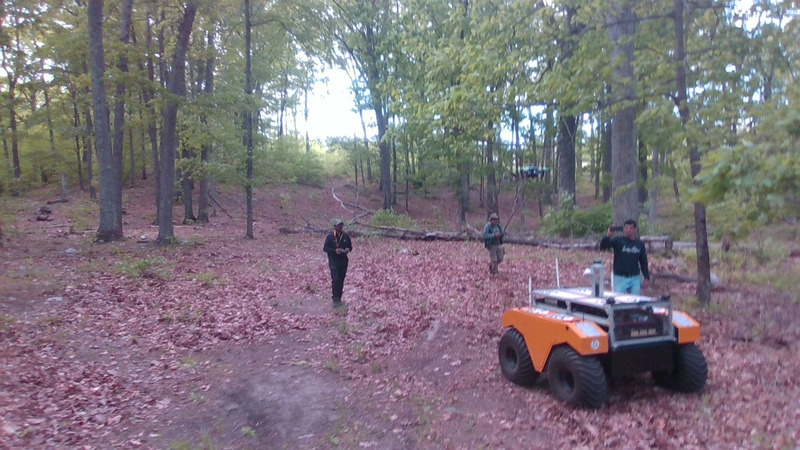}
     \end{subfigure}
     \begin{subfigure}[b]{0.47\linewidth}
        \centering
        \includegraphics[width=\textwidth]{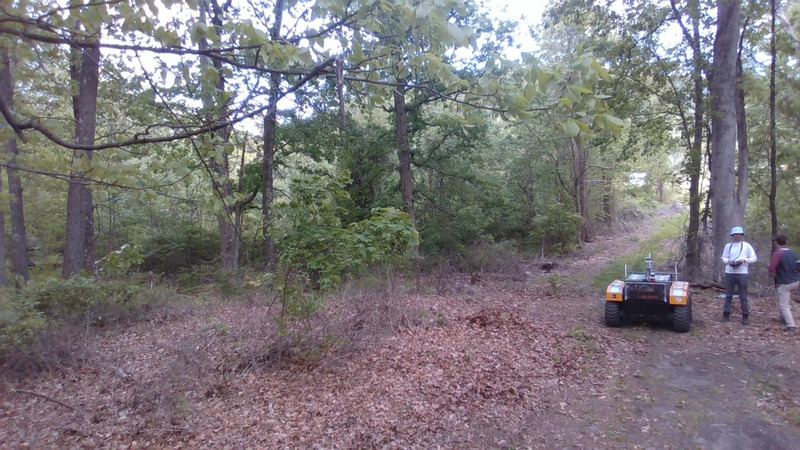}
    \end{subfigure}
    \caption{Outdoor-FOREST environment. Pointcloud data (top row) is captured by the ground robots, and the color images (bottom row) are captured by the aerial robots. This snapshot captures the scenarios when two groups of robot subteams encounter each other.}
    \label{fig:samples-FOREST}
    \vspace{-10pt}
\end{figure}
\begin{table}[t]
    \centering
    \caption{Sequence Info}
    \begin{tabular}{c c c c c c c }
        \toprule 
        Sequence & Scene & Time & RGB Frames & Instances/Frame \\
        \midrule 
        NYUARPL & Indoor & $400$ \si{s} & $48000$ & $9$\\
        \midrule
        CHAIR-1 & Indoor & $180$ \si{s} & $10800$ & $11$\\
        \midrule
        HALLWAY & Indoor & $200$ \si{s} & $12000$ & $3$\\
        \midrule
        \midrule
        HOUSEA & Outdoor & $440$ \si{s} & $52800$ & $8$\\
        \midrule
        HOUSEB & Outdoor & $440$ \si{s} & $52800$ & $9$\\
        \midrule
        FOREST & Outdoor & $225$ \si{s} & $27000$ & $5$\\
        \bottomrule 
    \end{tabular}
    \label{tab:sequences}
        \vspace{-20pt}
\end{table}

\subsection{Outoor Sequences}
In the outdoor scenario, we use two Warthog ground robots and two aerial robots. We present three sequences with two Warthog ground robots and two aerial robots, and two sequences with one Warthog ground robot and one aerial robot. Sample dataset qualitative results are shown in Fig.~\ref{fig:samples-FOREST}.
The outdoor sequences are captured in two outdoor environments. The first environment, named 'HOUSE', is an open space of $150~\si{m} \times 30~\si{m}$  situated by a forest, encompassing houses, roads, trees, and rocks. The houses in this setting can be used as object detection targets, thus providing a practical context for target detection tasks. 
The second environment, 'FOREST', is a space of $100~\si{m} \times 80~\si{m}$ nestled within the forest, presenting a rich composition of trees, ground variations, and rocks. These environments introduce various features and obstacles commonly encountered in outdoor environments, thus offering a robust platform for training and testing algorithms. 
In all outdoor sequences, the robot team splits up into two subteams each composed of one aerial robot and one ground robot. The aerial robots fly above the ground robots for most of the time, and detach from the ground robot to explore the area where ground robots have difficulty maneuvering.  The aerial robots fly at a height ranging from $2.0~\si{m}$ to $10.0~\si{m}$. The ground robots drive at a speed up to $1.5~\si{m/s}$.
In 'Outdoor-HOUSEB', we also experiment with switching the pairing of the aerial and ground robots in the middle of the sequence to study the effect of formation on collaborative perception performances.

\begin{figure}[b]
    \centering\includegraphics[width=0.45\linewidth, height=0.30\linewidth]{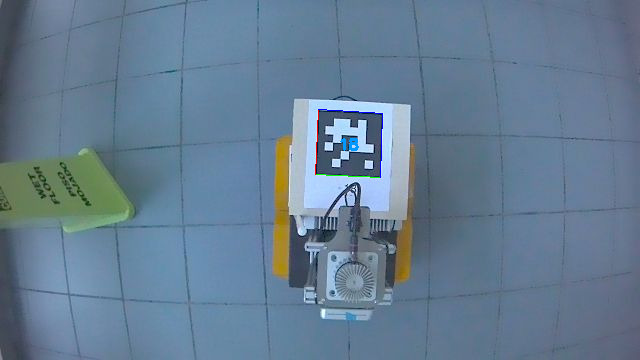}
    \includegraphics[width=0.45\linewidth, height=0.30\linewidth]{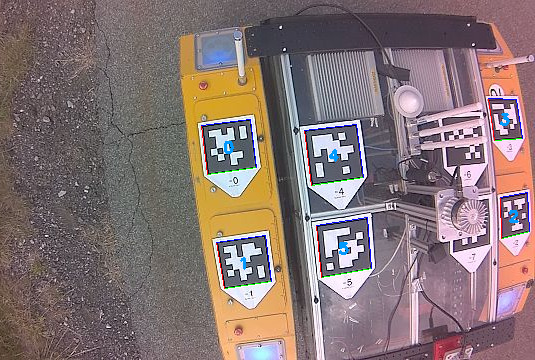}
    \caption{Detected AprilTag mounted on the ground robots captured from one aerial robot's downward-facing camera.}
    \label{fig:downward-apriltag}
\end{figure}

\begin{figure*}[t]
    \centering
    \rotatebox{90}{\parbox{0.13\linewidth}{\centering \scriptsize Semantics}}
    \begin{subfigure}[b]{0.24\linewidth}
         \centering
         \includegraphics[height=0.5510\textwidth, width=\textwidth]{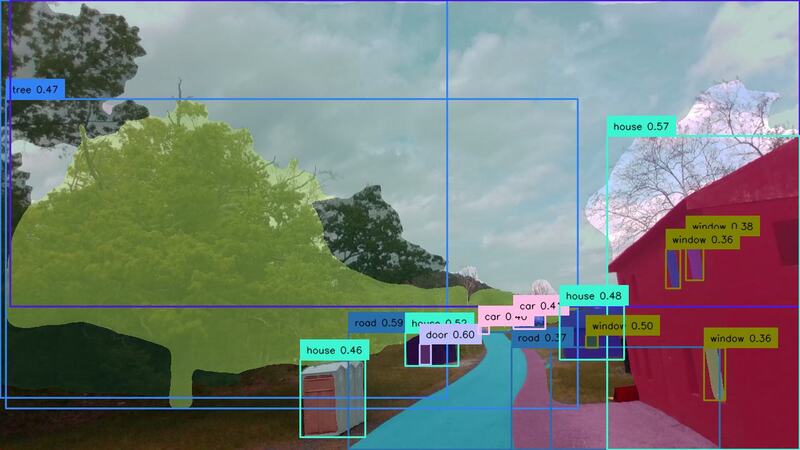}
     \end{subfigure}
     \begin{subfigure}[b]{0.24\linewidth}
         \centering
         \includegraphics[height=0.5510\textwidth, width=\textwidth]{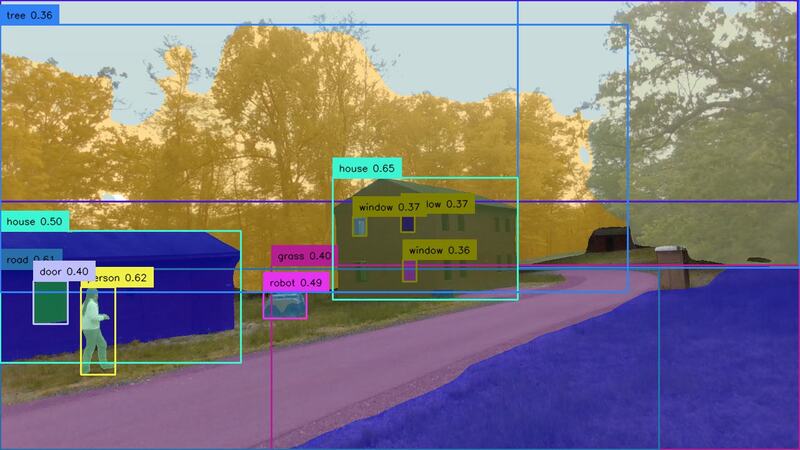}
     \end{subfigure}
     \begin{subfigure}[b]{0.24\linewidth}
         \centering
         \includegraphics[height=0.5510\textwidth, width=\textwidth]{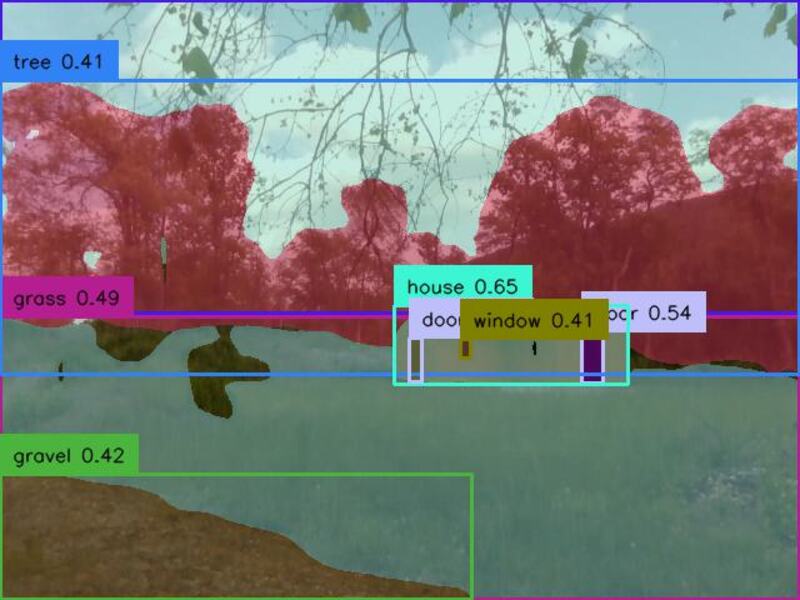}
     \end{subfigure}
      \begin{subfigure}[b]{0.24\linewidth}
         \centering
         \includegraphics[height=0.5510\textwidth, width=\textwidth]{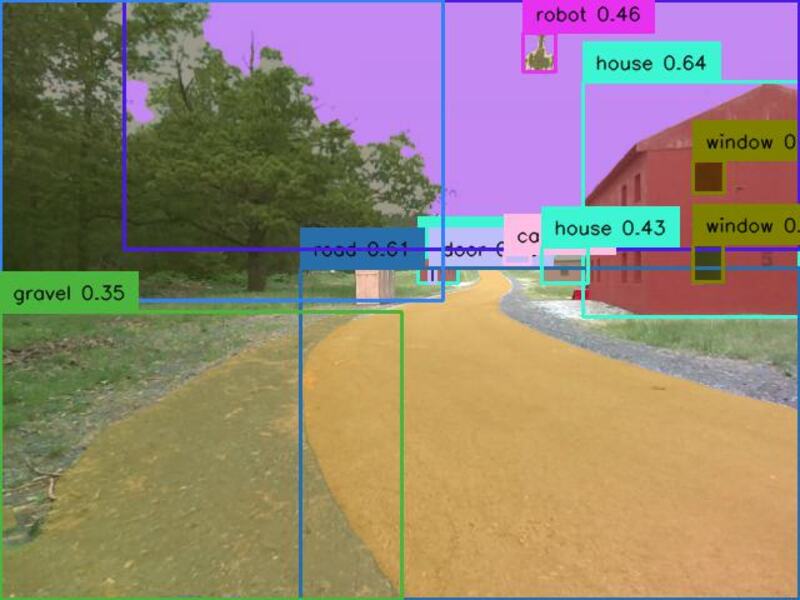}
     \end{subfigure}
\rotatebox{90}{\parbox{0.13\linewidth}{\centering \scriptsize Depth}}
     \begin{subfigure}[b]{0.24\linewidth}
         \centering
         \includegraphics[height=0.5510\textwidth, width=\textwidth]{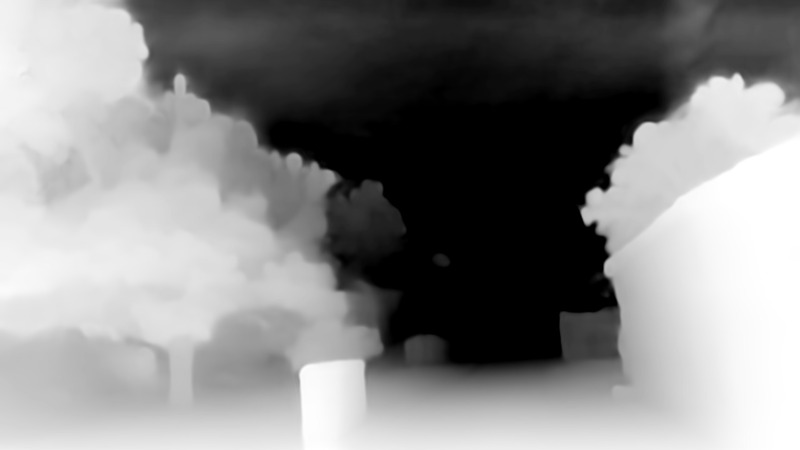}
     \end{subfigure}
     \begin{subfigure}[b]{0.24\linewidth}
         \centering
         \includegraphics[height=0.5510\textwidth, width=\textwidth]{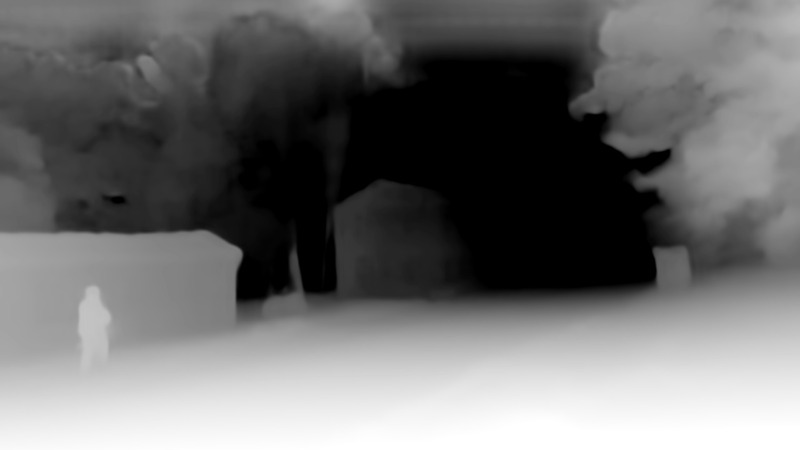}
     \end{subfigure}
     \begin{subfigure}[b]{0.24\linewidth}
         \centering
         \includegraphics[height=0.5510\textwidth, width=\textwidth]{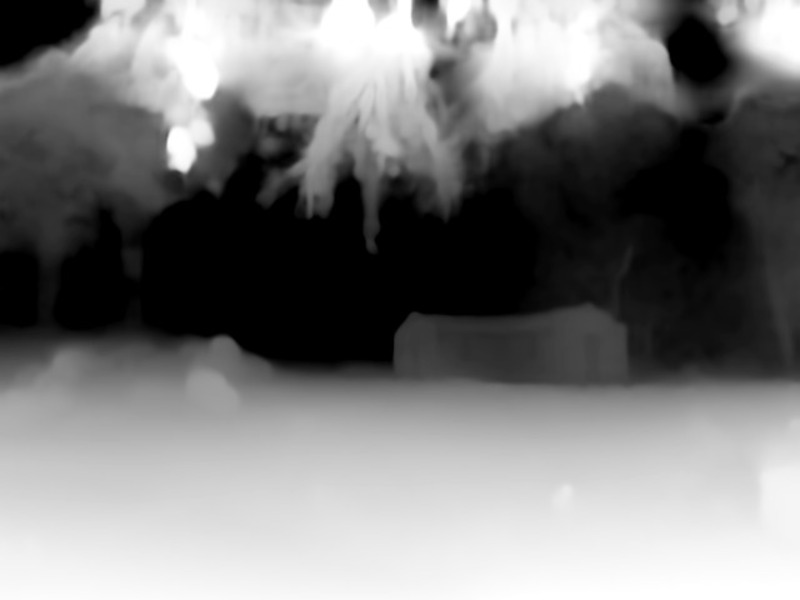}
     \end{subfigure}
      \begin{subfigure}[b]{0.24\linewidth}
         \centering
         \includegraphics[height=0.5510\textwidth, width=\textwidth]{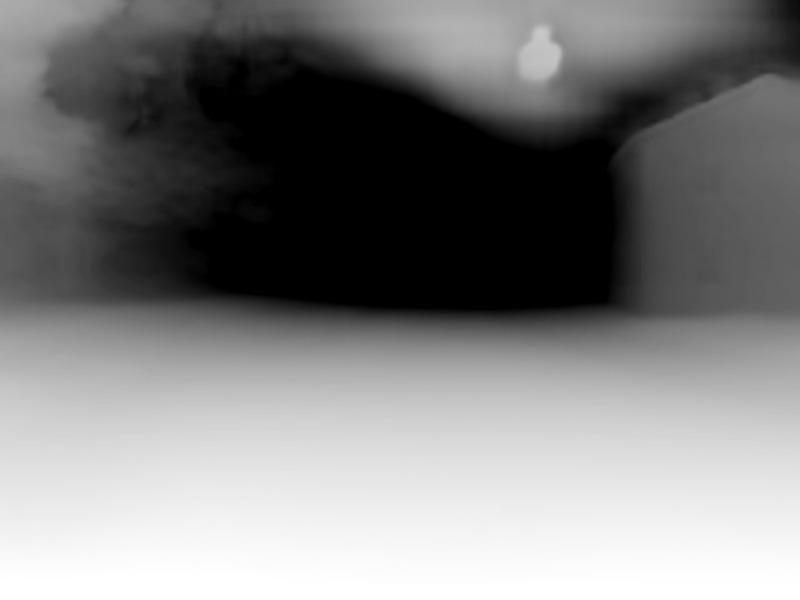}
     \end{subfigure}
\rotatebox{90}{\parbox{0.13\linewidth}{\centering \scriptsize Semantics}}
     \begin{subfigure}[b]{0.24\linewidth}
         \centering
         \includegraphics[height=0.5510\textwidth, width=\textwidth]{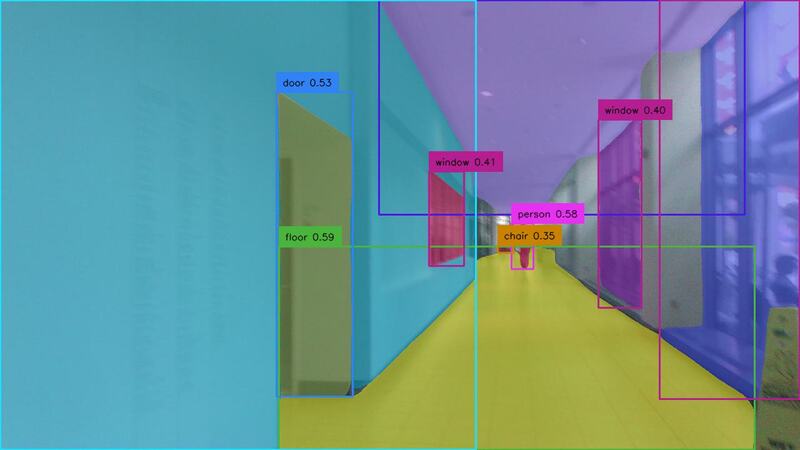}
     \end{subfigure}
     \begin{subfigure}[b]{0.24\linewidth}
         \centering
         \includegraphics[height=0.5510\textwidth, width=\textwidth]{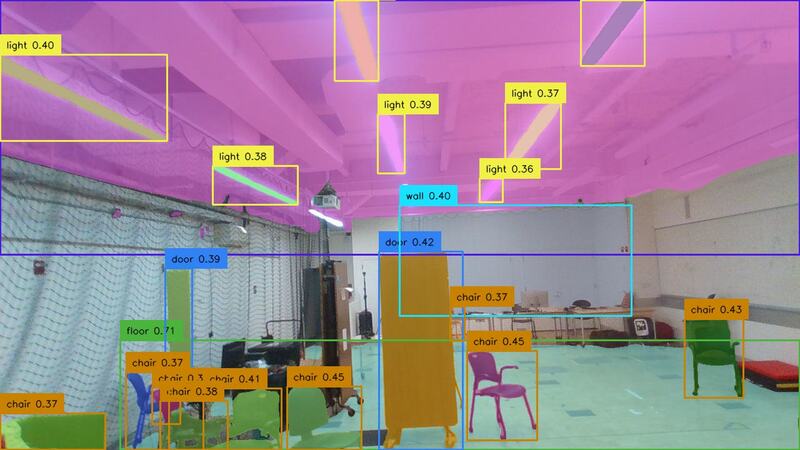}
     \end{subfigure}
     \begin{subfigure}[b]{0.24\linewidth}
         \centering
         \includegraphics[height=0.5510\textwidth, width=\textwidth]{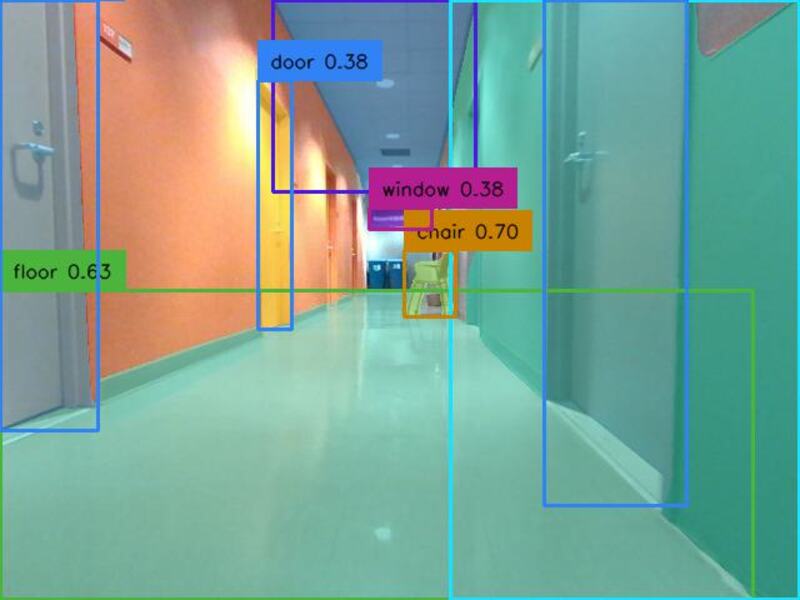}
     \end{subfigure}
      \begin{subfigure}[b]{0.24\linewidth}
         \centering
         \includegraphics[height=0.5510\textwidth, width=\textwidth]{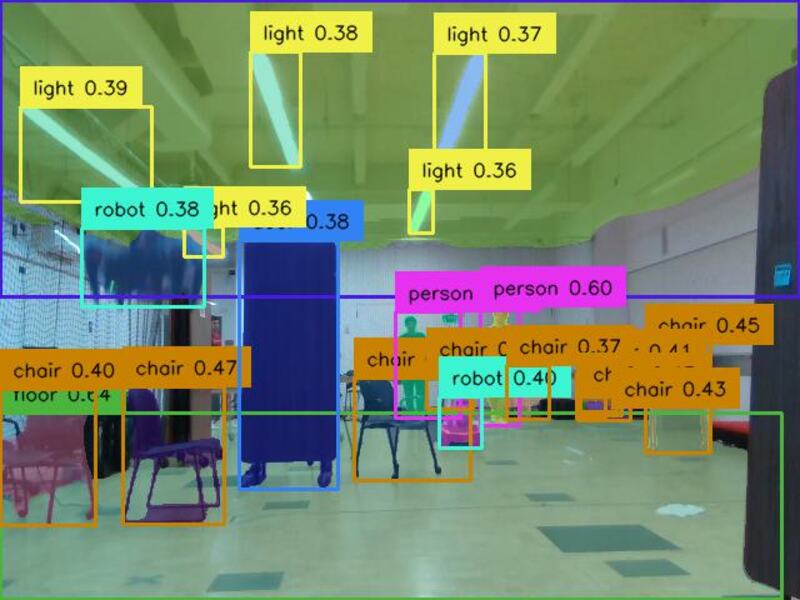}
     \end{subfigure}
\rotatebox{90}{\parbox{0.13\linewidth}{\centering \scriptsize Depth}}
     \begin{subfigure}[b]{0.24\linewidth}
         \centering
         \includegraphics[height=0.5510\textwidth, width=\textwidth]{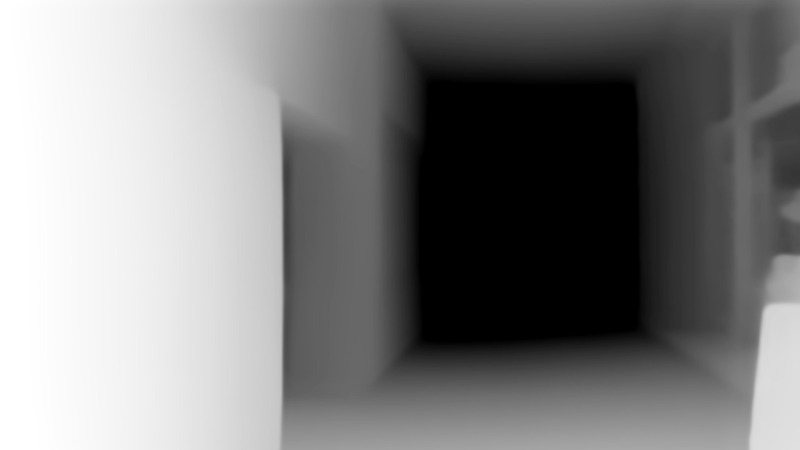}
     \end{subfigure}
     \begin{subfigure}[b]{0.24\linewidth}
         \centering
         \includegraphics[height=0.5510\textwidth, width=\textwidth]{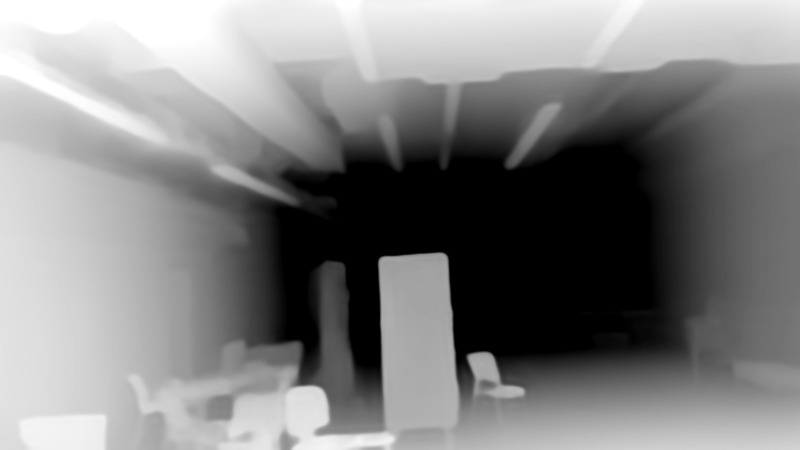}
     \end{subfigure}
     \begin{subfigure}[b]{0.24\linewidth}
         \centering
         \includegraphics[height=0.5510\textwidth, width=\textwidth]{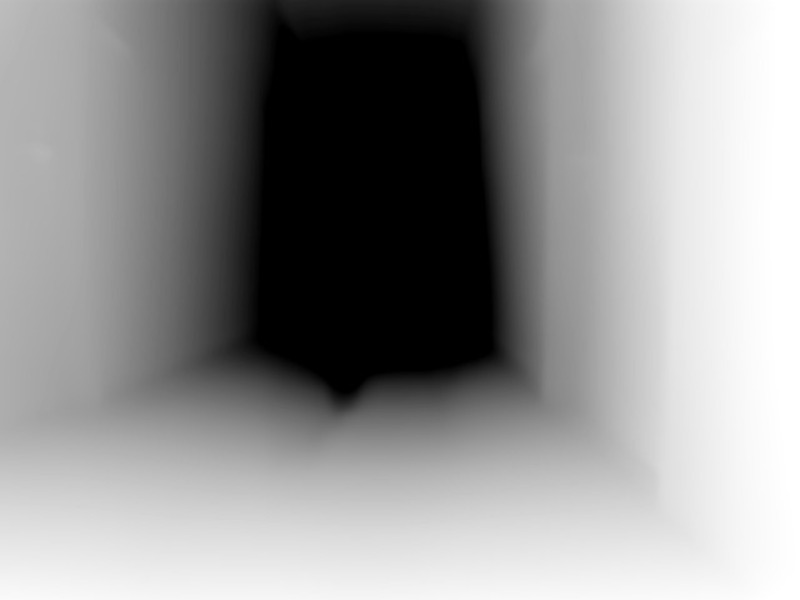}
     \end{subfigure}
      \begin{subfigure}[b]{0.24\linewidth}
         \centering
         \includegraphics[height=0.5510\textwidth, width=\textwidth]{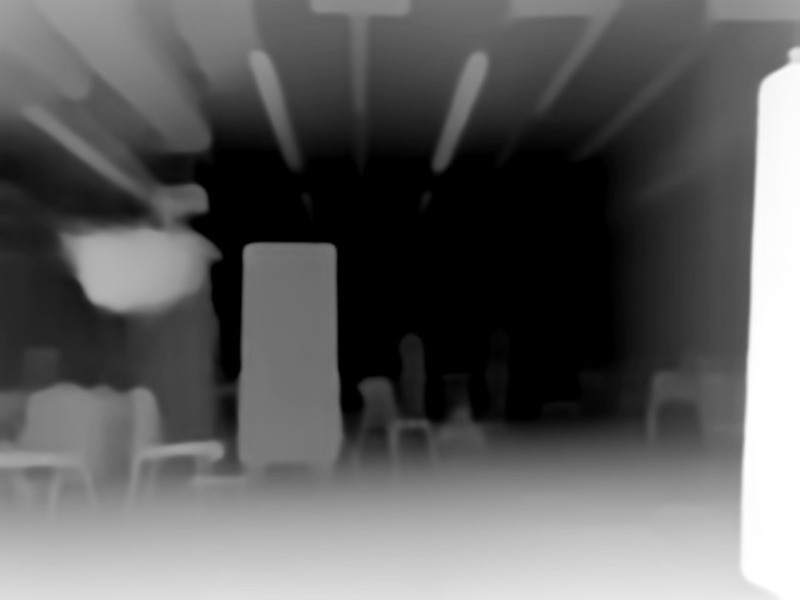}
     \end{subfigure}
     \begin{subfigure}[b]{0.48\linewidth}
     \caption{Aerial Robot}
     \end{subfigure}
     \begin{subfigure}[b]{0.48\linewidth}
     \caption{Ground Robot}
     \end{subfigure}
    \caption{Zero-shot annotation of semantics and depth in indoor and outdoor environments from aerial and ground robots.}
    \label{fig:zeroshot-annotation}
    \vspace{-10pt}
\end{figure*}

\section{Calibration}~\label{sec:calibration}
\textbf{Aerial Robot Calibration}.
The parameters of the aerial robot sensors are calibrated using the Kalibr toolbox~\cite{rehder2016extending}. The calibration process includes the intrinsic calibration~\cite{zhang2000flexible} of the RGB and RGBD cameras respectively, the extrinsic calibration~\cite{furgale2013unified} of the RGBD camera and IMU, and the extrinsic calibration of the RGB camera and IMU. We also calibrate the IMU intrinsic parameter by conducting IMU Allan covariance analysis.

\textbf{Ground Robot Calibration}. 
We employ multi-sensor Graph-based Calibration to produce a unified graph-based representation of the ground robots' sensor nodes~\cite{owens2015msg}. These sensors include LiDAR, RGBD, and stereo cameras from Table~\ref{tab:groundrobotsensors}. The Ouster LiDAR (OS1) is chosen as the root/reference frame in the robot's transform hierarchy. The graph optimization approach computes pairwise relative poses for each sensor node.

\textbf{Air-Ground Calibration}.
To ensure the accuracy and consistency of our data, we implemented a meticulous air-ground system sensor calibration process. 
For extrinsic calibration of the start position of the robot team, a calibration board is positioned in front of all the forward-facing cameras of the robot team, which ensures the spatial alignment of different modalities and data consistency, therefore providing a reliable foundation for subsequent research.

\section{Data Annotation}~\label{sec:data} 
\vspace{-20pt}
\subsection{Pose Estimation}
We employ an individual state estimation system for each robot. For aerial robots, the fusion of Global Positioning System (GPS) data with Stereo Visual Inertial Odometry provides accurate pose estimation. In our autonomy stack, we adopt OpenVINS~\cite{geneva2020openvins} for stereo Visual Inertial Odometry (VIO). Conversely, the ground robots utilize IMU and wheel odometry for reliable state estimation. The maximum speed of the aerial robot is $2.1~\si{m/s}$, the maximum speed of the ground robot is $1.75~\si{m/s}$.

Moreover, computing the cross-robot relative pose estimation adds an additional level of precision. This is achieved by identifying a bundle of AprilTags~\cite{olson2011apriltag}—fiducial marker systems commonly used in robotics—mounted on the ground robots from the vantage point of the aerial robots, illustrated in Fig.~\ref{fig:downward-apriltag}.
We calibrate the extrinsic between each tag within the AprilTab bundle by capturing a sequence of images from a camera with known intrinsic parameters. Therefore, we can calculate the relative transformation between the AprilTag bundle frame with respect to the camera frame. 
\subsection{Zero-shot Semantics and Depth Annotation}
In addition to pose estimation, we provide semantic and depth annotations within our dataset. We harness the power of foundation models to deliver zero-shot semantic and depth annotation. For semantics annotation, we adopt the combination of RAM~\cite{zhang2023recognize}, Grounding DINO~\cite{liu2023grounding} and SAM~\cite{kirillov2023segment} to enable the acquisition of 2D instance segmentation masks directly from the camera sensors. We obtain 2D bounding boxes by post-processing the mask annotations. For depth estimation, we adopt Zoedepth~\cite{bhat2023zoedepth} to produce zero-shot monocular depth estimation annotation. 
Due to the limited baseline of Intel Realsense stereo camera, the quality of traditional stereo depth matching such as SGBM~\cite{hirschmuller2005accurate} and graph-cut stereo matching~\cite{kolmogorov2014kolmogorov} are worse than state-of-the-art monocular depth estimation such as Zoedepth, which combines relative and metric depth. We cannot apply learning-based stereo-matching algorithms since these are designed for color rather than grayscale cameras which are employed on our robots. 
For semantics annotation, we utilize temporal propagation among multiple frames to increase the temporal consistency of the zero-shot prediction. Temporal consistent annotation is essential to denoise the annotation and it enables video-based dense prediction tasks. Specifically, we leverage \cite{yang2023track} as temporal propagation model to propagate the segments on keyframes to consecutive frames. The employment of zero-shot learning, wherein the model is not explicitly trained on task-specific data, increases the generalization and applicability of the annotations. Therefore, they enhance the usability of our dataset, facilitating robust multi-robot collaborative perception research. Sample qualitative results are illustrated in Fig.~\ref{fig:zeroshot-annotation}.

\begin{figure*}[t]
    \centering
    \rotatebox{90}{\parbox{0.09\linewidth}{\centering \scriptsize RGB}}
     \begin{subfigure}[b]{0.157\linewidth}
         \centering
         \includegraphics[width=\textwidth]{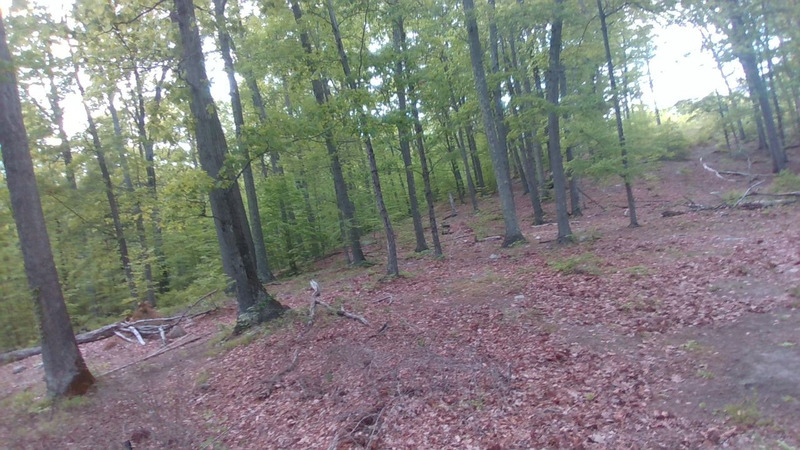}
     \end{subfigure}
     \begin{subfigure}[b]{0.157\linewidth}
         \centering
         \includegraphics[width=\textwidth]{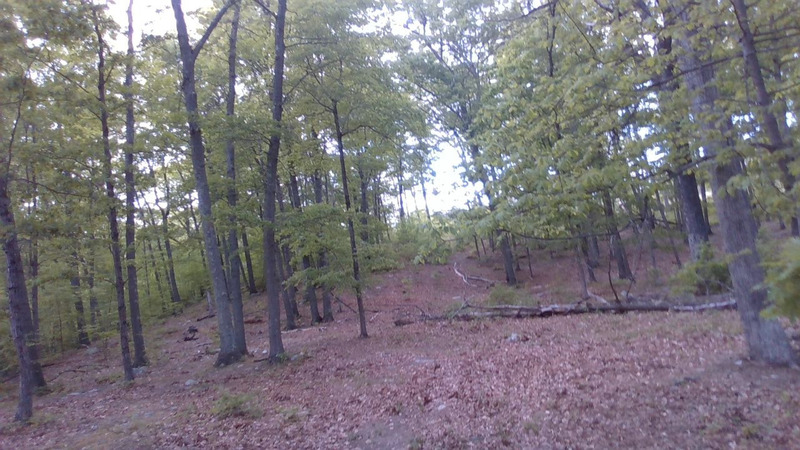}
     \end{subfigure}
     \begin{subfigure}[b]{0.157\linewidth}
         \centering
         \includegraphics[width=\textwidth, height=0.5625\textwidth]{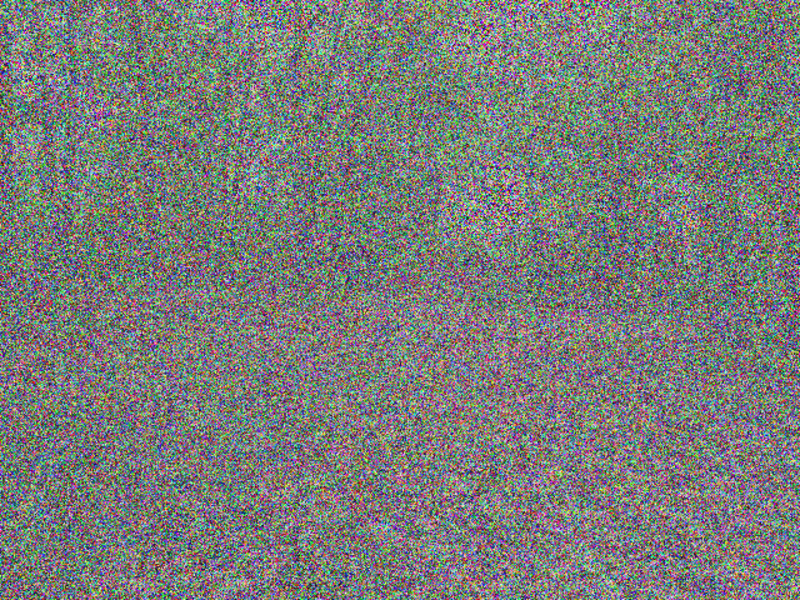}
     \end{subfigure}
     \begin{subfigure}[b]{0.157\linewidth}
         \centering
         \includegraphics[width=\textwidth]{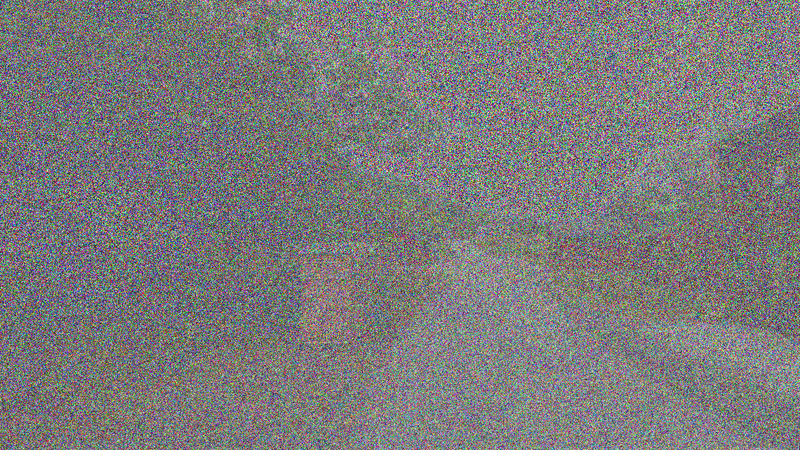}
     \end{subfigure}
\begin{subfigure}[b]{0.157\linewidth}
         \centering
         \includegraphics[width=\textwidth]{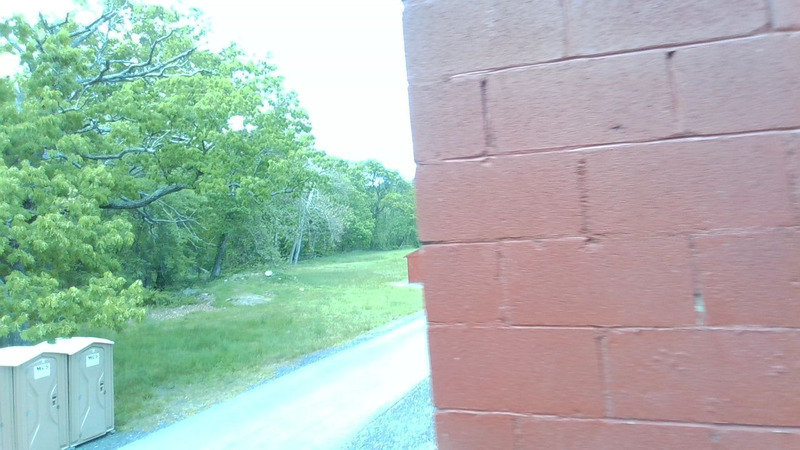}
     \end{subfigure}
     \begin{subfigure}[b]{0.157\linewidth}
         \centering
         \includegraphics[width=\textwidth, height=0.5625\textwidth]{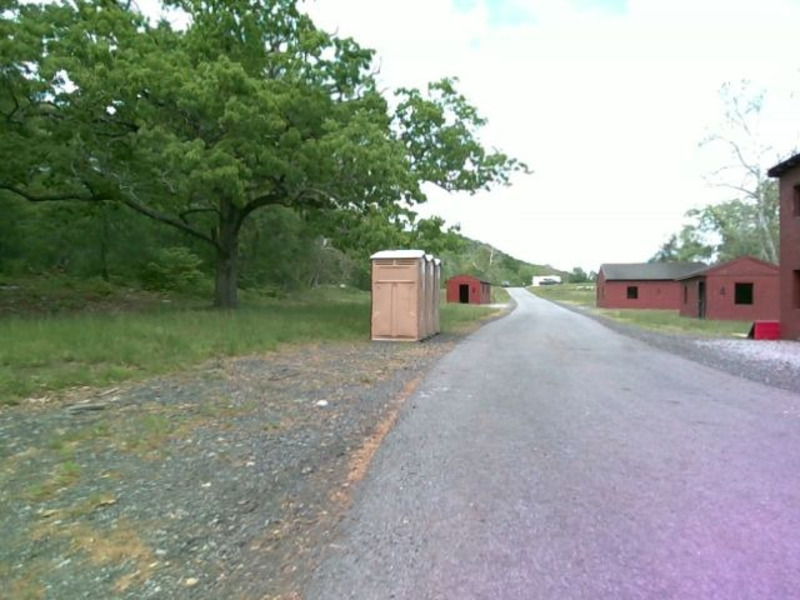}
     \end{subfigure}
\rotatebox{90}{\parbox{0.09\linewidth}{\centering \scriptsize GroundTruth}}
     \begin{subfigure}[b]{0.157\linewidth}
         \centering
         \includegraphics[width=\textwidth]{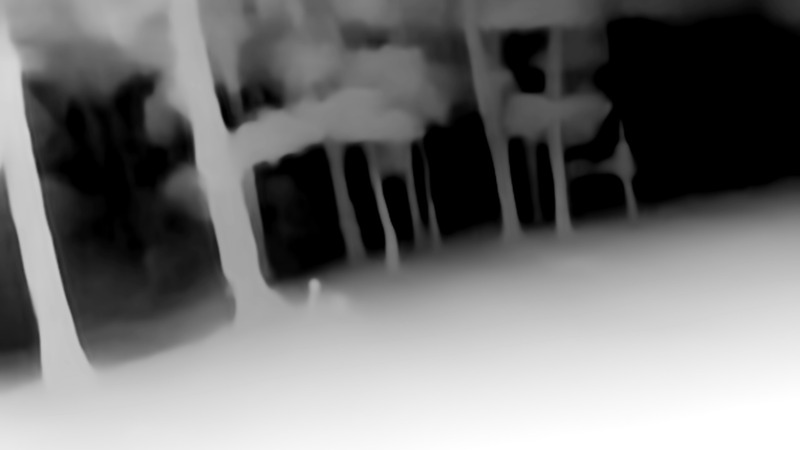}
     \end{subfigure}
     \begin{subfigure}[b]{0.157\linewidth}
         \centering
         \includegraphics[width=\textwidth]{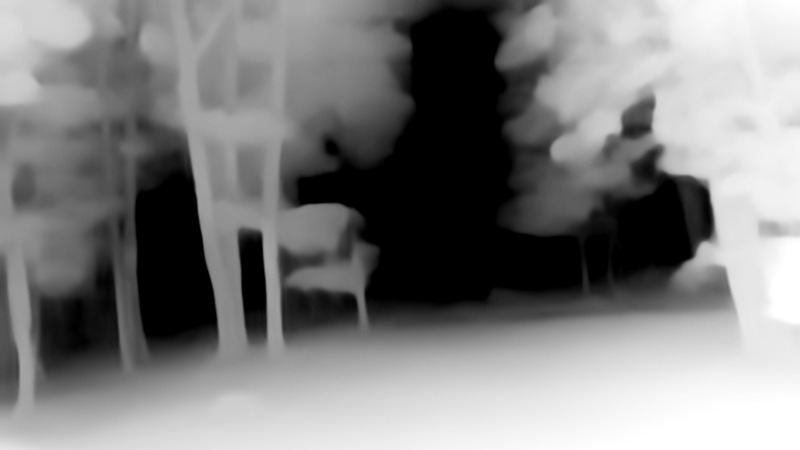}
     \end{subfigure}
     \begin{subfigure}[b]{0.157\linewidth}
         \centering
         \includegraphics[width=\textwidth, height=0.5625\textwidth]{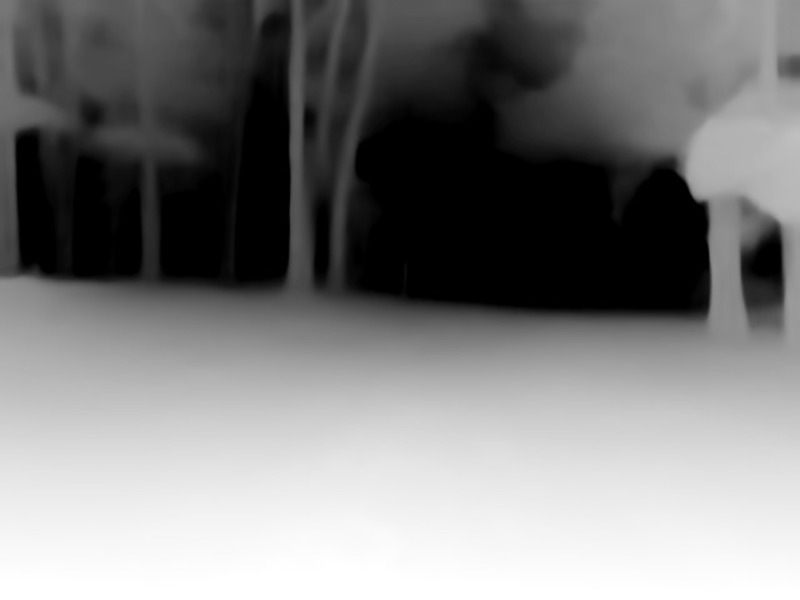}
     \end{subfigure}
     \begin{subfigure}[b]{0.157\linewidth}
         \centering
         \includegraphics[width=\textwidth]{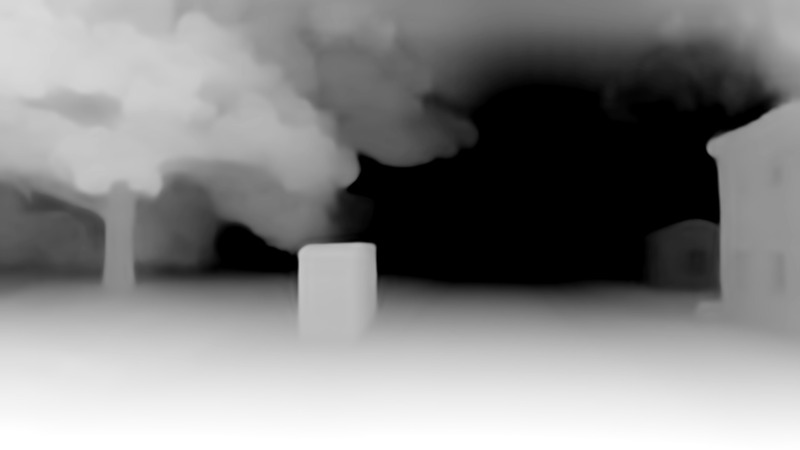}
     \end{subfigure}
     \begin{subfigure}[b]{0.157\linewidth}
         \centering
         \includegraphics[width=\textwidth]{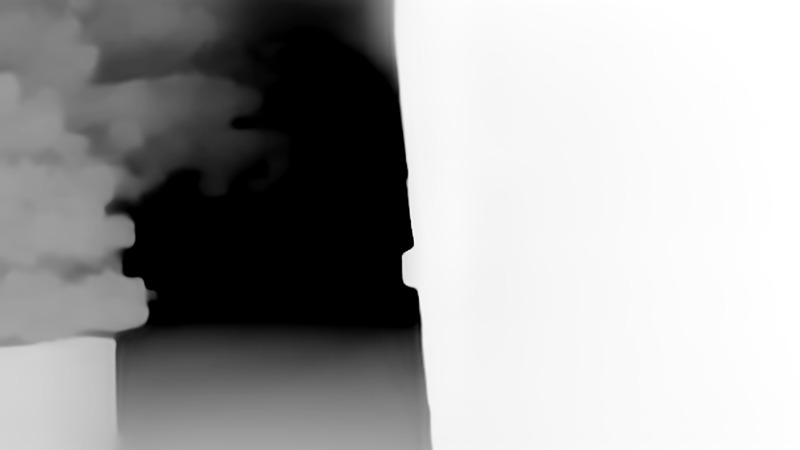}
     \end{subfigure}
     \begin{subfigure}[b]{0.157\linewidth}
         \centering
         \includegraphics[width=\textwidth, height=0.5625\textwidth]{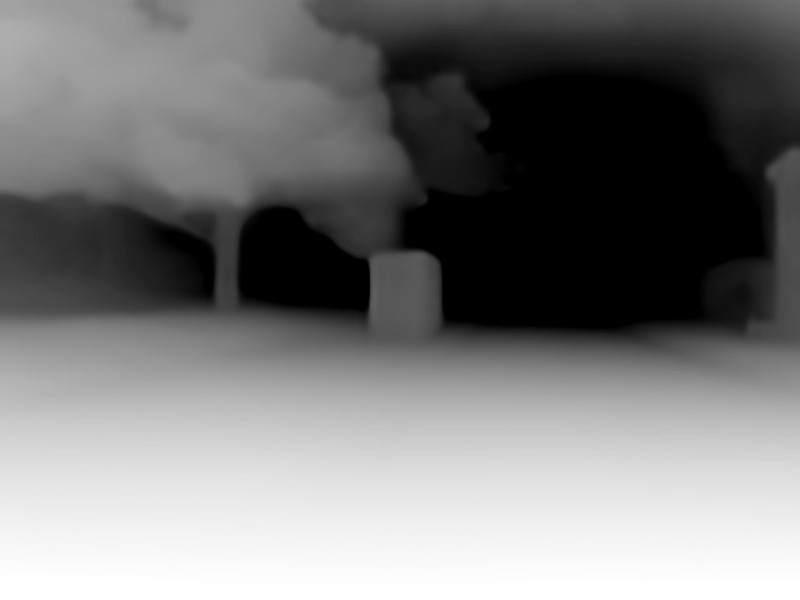}
     \end{subfigure}
\rotatebox{90}{\parbox{0.09\linewidth}{\centering \scriptsize Baseline}}
     \begin{subfigure}[b]{0.157\linewidth}
         \centering
         \includegraphics[width=\textwidth]{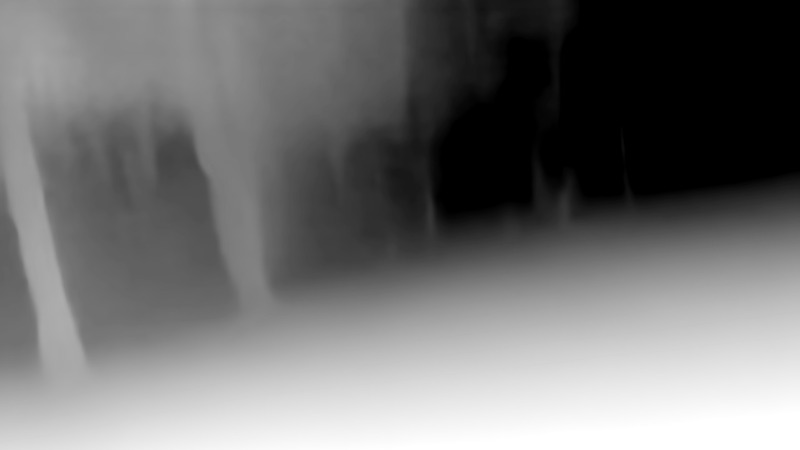}
     \end{subfigure}
     \begin{subfigure}[b]{0.157\linewidth}
         \centering
         \includegraphics[width=\textwidth]{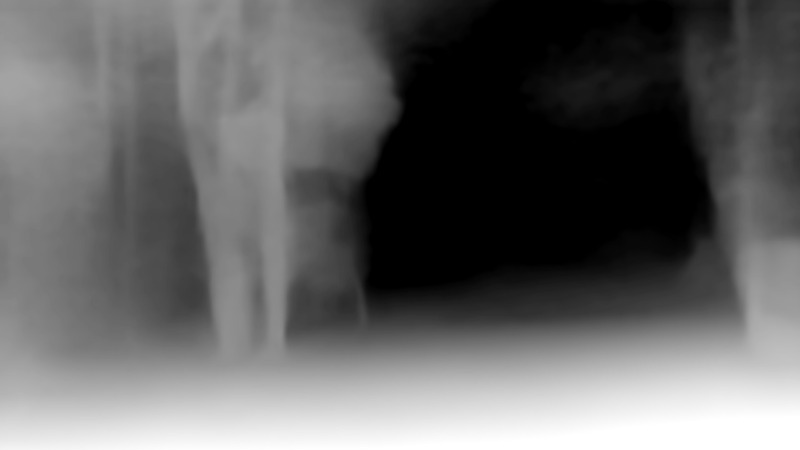}
     \end{subfigure}
     \begin{subfigure}[b]{0.157\linewidth}
         \centering
         \includegraphics[width=\textwidth, height=0.5625\textwidth]{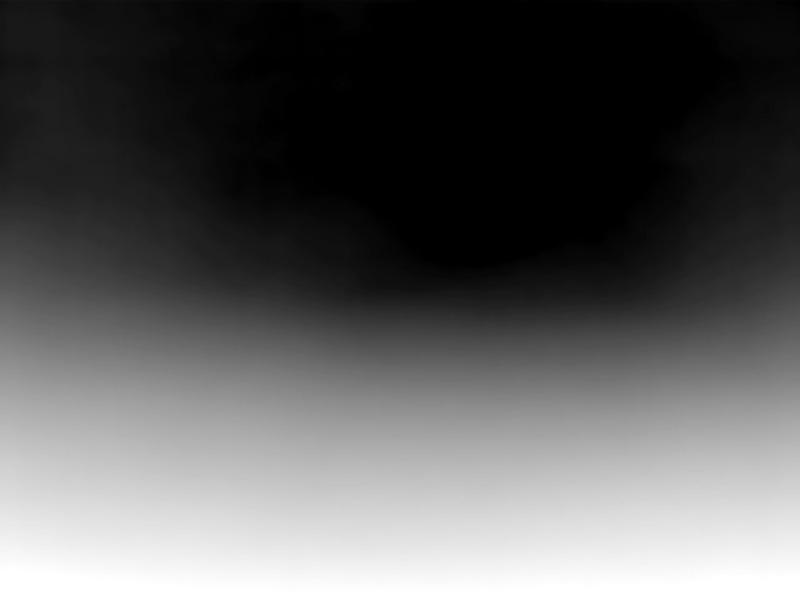}
     \end{subfigure}
     \begin{subfigure}[b]{0.157\linewidth}
         \centering
         \includegraphics[width=\textwidth]{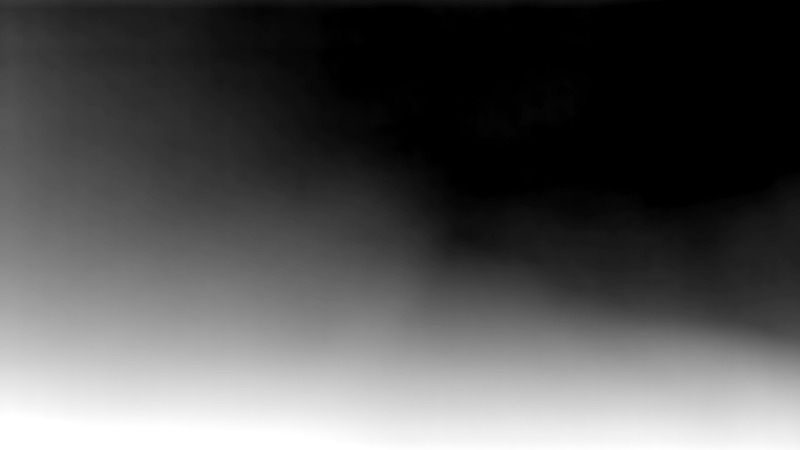}
     \end{subfigure}
     \begin{subfigure}[b]{0.157\linewidth}
         \centering
         \includegraphics[width=\textwidth]{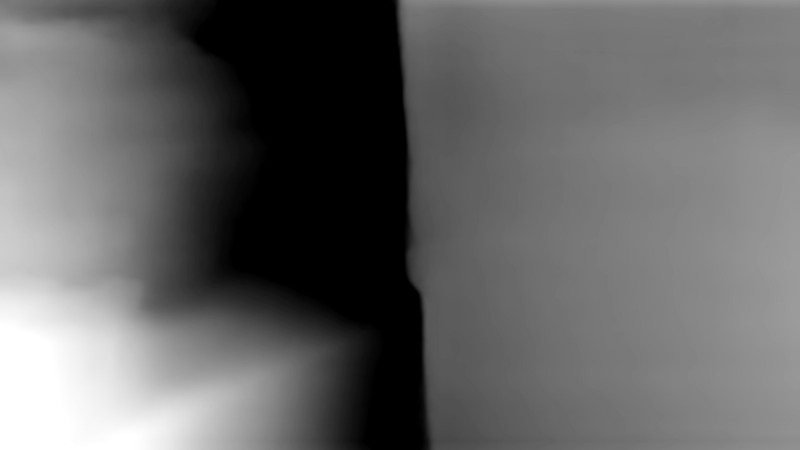}
     \end{subfigure}
     \begin{subfigure}[b]{0.157\linewidth}
         \centering
         \includegraphics[width=\textwidth, height=0.5625\textwidth]{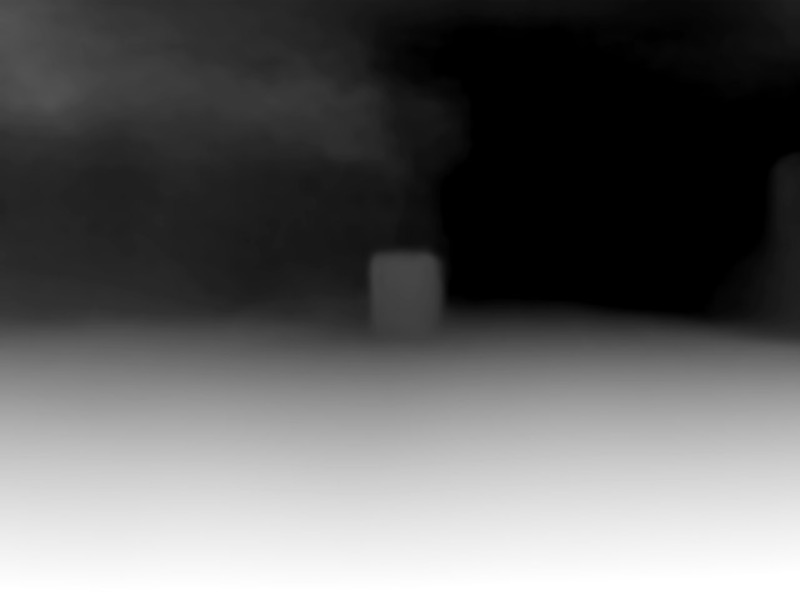}
     \end{subfigure}
\rotatebox{90}{\parbox{0.09\linewidth}{\centering \scriptsize MultiRobot}}
     \begin{subfigure}[b]{0.157\linewidth}
         \centering
         \includegraphics[width=\textwidth]{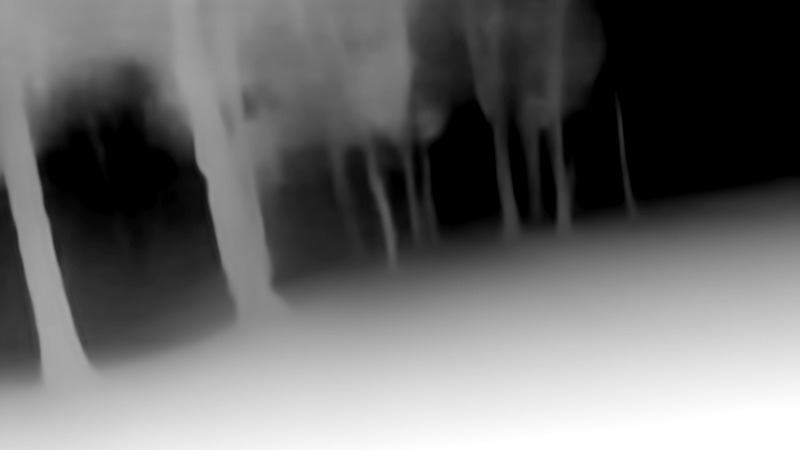}
     \end{subfigure}
     \begin{subfigure}[b]{0.157\linewidth}
         \centering
         \includegraphics[width=\textwidth]{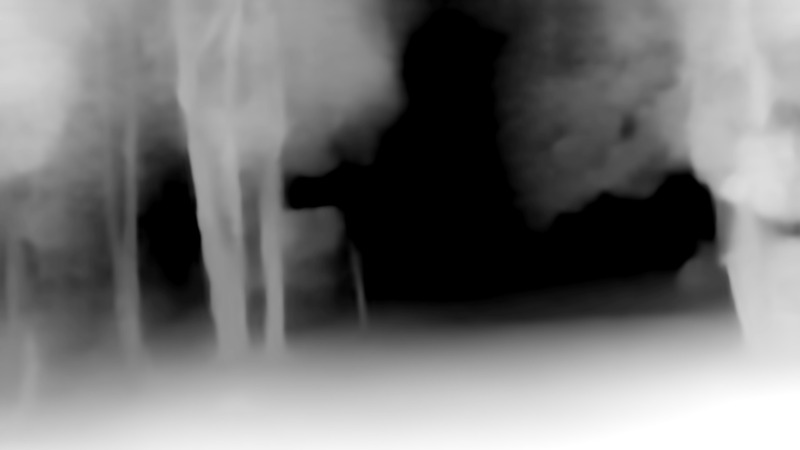}
     \end{subfigure}
     \begin{subfigure}[b]{0.157\linewidth}
         \centering
         \includegraphics[width=\textwidth, height=0.5625\textwidth]{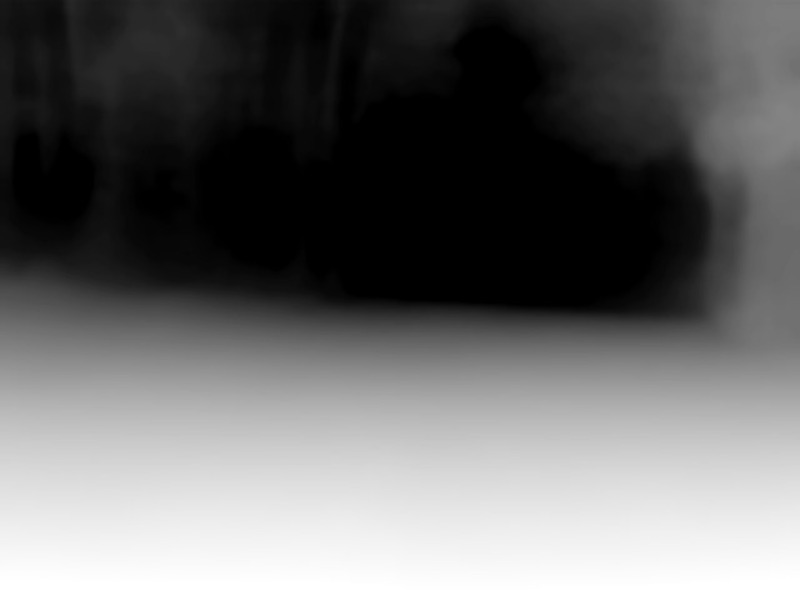}
     \end{subfigure}
     \begin{subfigure}[b]{0.157\linewidth}
         \centering
         \includegraphics[width=\textwidth]{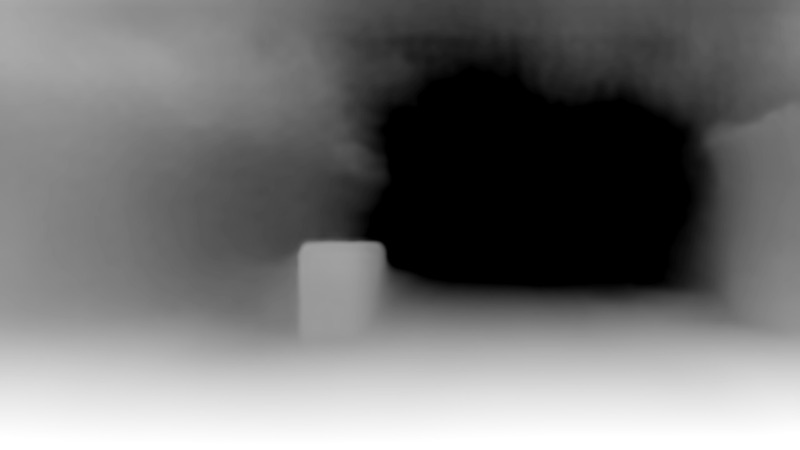}
     \end{subfigure}
     \begin{subfigure}[b]{0.157\linewidth}
         \centering
         \includegraphics[width=\textwidth]{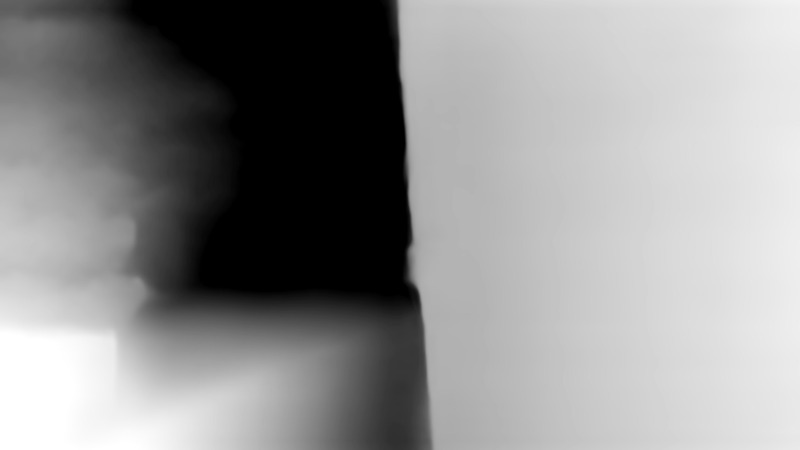}
     \end{subfigure}
     \begin{subfigure}[b]{0.157\linewidth}
         \centering
         \includegraphics[width=\textwidth, height=0.5625\textwidth]{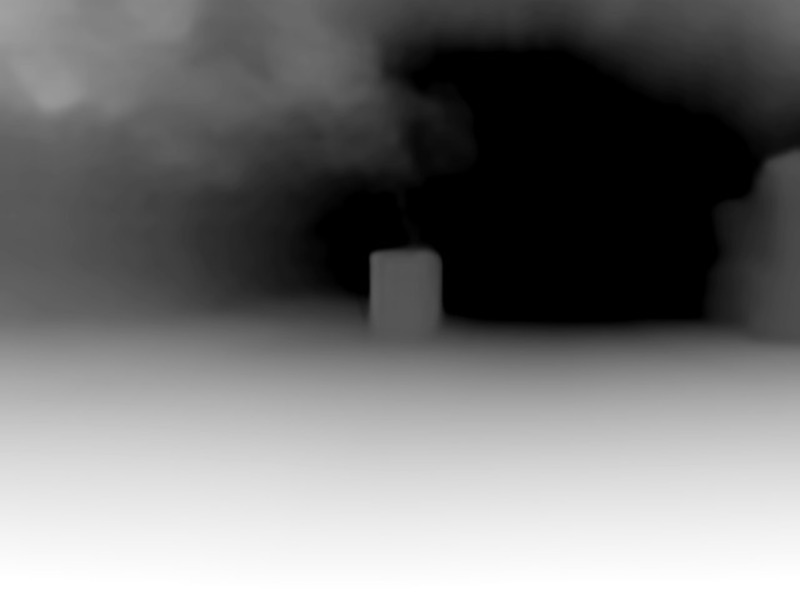}
     \end{subfigure}
     \begin{subfigure}[b]{0.157\linewidth}
     \caption{T1: Aerial}
     \end{subfigure}
     \begin{subfigure}[b]{0.157\linewidth}
     \caption{T1: Aerial}
     \end{subfigure}
     \begin{subfigure}[b]{0.157\linewidth}
     \caption{T1: Ground}
     \end{subfigure}
     \begin{subfigure}[b]{0.157\linewidth}
     \caption{T2: Aerial}
     \end{subfigure}
     \begin{subfigure}[b]{0.157\linewidth}
     \caption{T2: Aerial}
     \end{subfigure}
     \begin{subfigure}[b]{0.157\linewidth}
     \caption{T2: Ground}
     \end{subfigure}
    \caption{Multi-robot collaborative depth estimation with two aerial robots and one ground robot. (a-c) and (d-f) show two scenarios (T1 and T2) with (c) and (d) robot image sensors corrupted The rows correspond to RGB images, groundtruth depth, single-robot baseline, and multi-robot collaboration results respectively.}\label{fig:qualitative_result}
    \vspace{-10pt}
\end{figure*}
\begin{figure}[t]
    \centering
    \rotatebox{90}{\parbox{0.18\linewidth}{\centering \scriptsize RGB}}
     \begin{subfigure}[b]{0.31\linewidth}
         \centering
         \includegraphics[width=\textwidth]{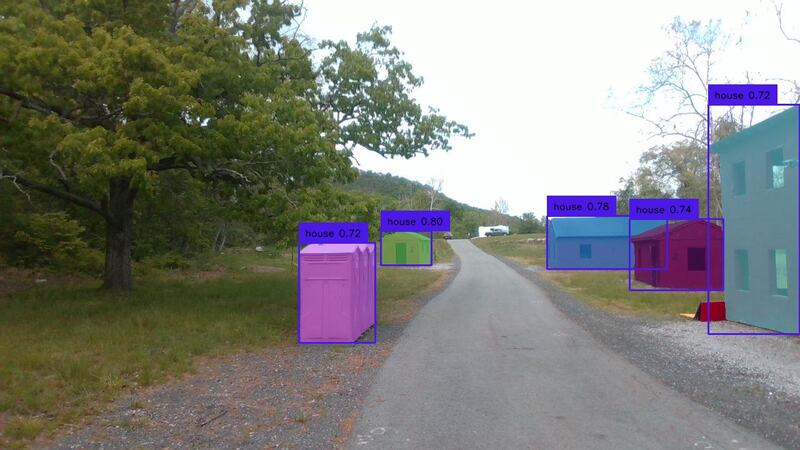}
     \end{subfigure}
\begin{subfigure}[b]{0.31\linewidth}
         \centering
         \includegraphics[width=\textwidth]{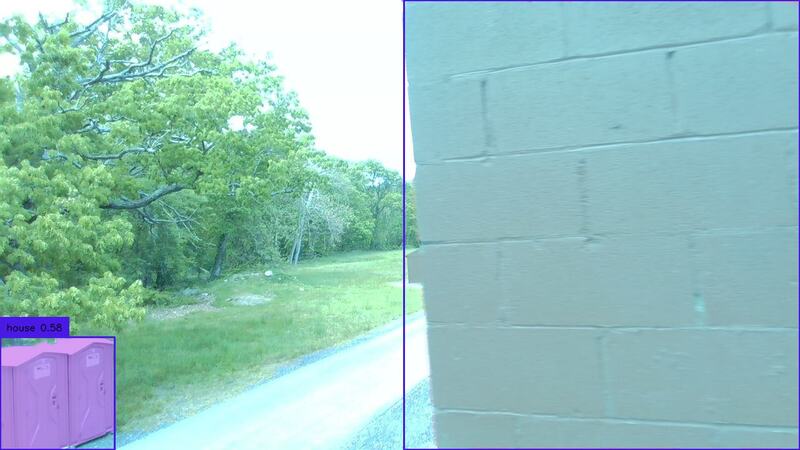}
     \end{subfigure}
     \begin{subfigure}[b]{0.31\linewidth}
         \centering
         \includegraphics[width=\textwidth, height=0.5625\textwidth]{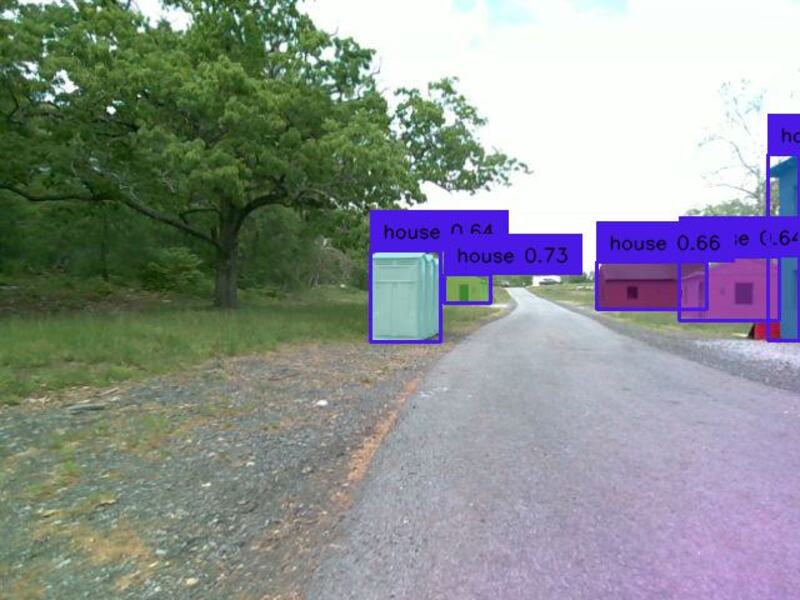}
     \end{subfigure}
     \rotatebox{90}{\parbox{0.18\linewidth}{\centering \scriptsize Baseline}}
\begin{subfigure}[b]{0.31\linewidth}
         \centering
         \includegraphics[width=\textwidth]{fig/qualitative/b_race1_HOUSEA.jpg}
     \end{subfigure}
     \begin{subfigure}[b]{0.31\linewidth}
         \centering
         \includegraphics[width=\textwidth]{fig/qualitative/b_race5_HOUSEA.jpg}
     \end{subfigure}
     \begin{subfigure}[b]{0.31\linewidth}
         \centering
         \includegraphics[width=\textwidth, height=0.5625\textwidth]{fig/qualitative/b_wanda_HOUSEA.jpg}
     \end{subfigure}
     \rotatebox{90}{\parbox{0.18\linewidth}{\centering \scriptsize MultiRobot}}
\begin{subfigure}[b]{0.31\linewidth}
         \centering
         \includegraphics[width=\textwidth]{fig/qualitative/g_race1_HOUSEA.jpg}
     \end{subfigure}
     \begin{subfigure}[b]{0.31\linewidth}
         \centering
         \includegraphics[width=\textwidth]{fig/qualitative/g_race5_HOUSEA.jpg}
     \end{subfigure}
     \begin{subfigure}[b]{0.31\linewidth}
         \centering
         \includegraphics[width=\textwidth, height=0.5625\textwidth]{fig/qualitative/g_wanda_HOUSEA.jpg}
     \end{subfigure}
\begin{subfigure}[b]{0.31\linewidth}
     \caption{T2: Aerial}
     \end{subfigure}
     \begin{subfigure}[b]{0.31\linewidth}
     \caption{T2: Aerial}
     \end{subfigure}
     \begin{subfigure}[b]{0.31\linewidth}
     \caption{T2: Ground}
     \end{subfigure}
    \caption{Multi-robot collaborative semantics estimation with two aerial robots and one ground robot. Image sensor of (a) is corrupted. The rows correspond to groundtruth semantics, single-robot baseline, and multi-robot collaboration results.}\label{fig:qualitative_result_semantics}
    \vspace{-10pt}
\end{figure}

\section{Dataset Attributes}~\label{sec:attributes} 
The dataset presents several characteristics that are unique compared to other multi-robot datasets. These characteristics highlight real-world challenges and scenarios encountered when deploying a heterogeneous robot team in the wild. The statistics of the scene is illustrated in Tab.~\ref{tab:sequences}.
First, we present a mixed setup of indoor and outdoor scenarios. We showcase the transition between outdoor and indoor environments in 'Outdoor-HOUSEB' sequence using one aerial robot 'race5', this sequence is unique and challenging to perception algorithms, as it requires the robot to adapt to the drastic change in environment and sensor modalities. 
Second, for the multi-robot sequences, we provide continous spatio-temporal variations of the formation and relative position of the robot team memebers, introducing distinct and complementary viewpoint perspectives to guarantee a wide range of inter-robot interactions. This variety is crucial in studying how different formations and positional arrangements influence collaborative perception and team efficiency. 
In the indoor sequence 'Indoor-NYUARPL' and outdoor sequence 'Outdoor-FOREST', we split the robot team in two groups, one aerial robot and one ground robot in each group. This is a challenging setup for perception algorithms, as it requires the robot team to collaborate and share information to perform perception tasks under a wide range of overlap and occlusion conditions.
Third, the dataset preserves the real-world sensor noise and disturbances encountered when deploying a heterogeneous robot team in the wild. Cameras may experience color jittering, LiDARs can exhibit discontinuous intensities temporally, GPS may be denied or drifted in indoor or outdoor forest environments, IMU may have inconsistent intrinsic parameters because of temperature change, and clock synchronization can be lost because of the network delay. These real-world disturbances are crucial for perception algorithms to be robust and reliable.
Fourth, we also provide several single-robot sequences using one aerial robot exploring the 'HOUSE' and 'FOREST' environments. These sequences are captured using the same aerial robot as the multi-robot sequences, thus providing a consistent perspective for comparison. These sequences are useful to provide additional samples and annotations to facilitate perception research.
Finally, the availability of scenes where aerial robots detach from the ground robots when these cannot maneuver due to environmental constraints further shows the diverse capabilities of the proposed heterogeneous system and its usefulness for collaborative perception tasks.

\section{Use-Cases and Applications}~\label{sec:usecases}
Our dataset offers a unique platform for developing and evaluating collaborative perception algorithms in multi-robot systems. Researchers can explore the utilization of sensor data among various robots for complex multi-robot collaborative perception tasks including scenarios with sensor and environmental noise.  We showcase qualitative results of monocular depth estimation and semantics segmentation using Graph Neural Network~\cite{zhou2022multi} in Figs.~\ref{fig:qualitative_result} and~\ref{fig:qualitative_result_semantics}. Two aerial robots collaborate with one ground robot to overcome sensor noises by feature map communication. The results show that multi-robot collaboration improves the depth prediction and semantic segmentation robustness against single-agent baseline. The predictions of far-away objects are recovered, such as trees in depth estimation and houses in semantic segmentation.

Given that the dataset provides multi-modal sensor data from different types of robots, it can also be a tool for investigating sensor fusion techniques. This could lead to the development of more robust algorithms for multi-robot mapping~\cite{gao2023deep}, object detection~\cite{hu2022where2comm}, and decision-making~\cite{cui2022coopernaut} for real-world settings in varying indoor and outdoor conditions.

\section{Conclusion}~\label{sec:conclusion}
In this paper, we presented a comprehensive multi-robot, multi-modal, and multi-rate air-ground dataset captured in diverse real-world indoor and outdoor environments. The goal is to provide a substantial resource for the research community to investigate and advance the field of multi-robot collaborative perception.
Specifically, the dataset includes data from two types of robots (i.e., ground and aerial) equipped with multiple heterogeneous sensing modalities. A substantial effort was invested in covering multiple distinct indoor and outdoor environments, specifically 'HOUSE' and 'FOREST'. This involved varying navigation sequences and formation therefore ensuring to capure heterogeneous sensing data affected by real-world sensor noise and
disturbances and with large spatio-temporal distinct viewpoints variations and complementary perspectives. The data and annotations offer a unique opportunity to delve into several challenging research problems and aspects related to multi-robot systems, such as collaborative perception, sensor fusion, object detection, and formation control.
We believe that the limitations and challenges associated with our dataset will be a source of inspiration to the community to advance the research development of multi-robot perception systems and autonomy algorithms.

In the future, we plan to enrich the dataset with comprehensive 2D and 3D semantic annotations. This dataset should not be considered a monolithic solution. We will continue to expand it, drawing from additional tests in multiple additional environmental contexts and with diverse robot configurations. Feedback and suggestions from the community will be pivotal in guiding these enhancements. 
Finally, this dataset can serve to improve robotics simulation by real-to-sim.
It will serve as a stepping stone for future research in collaborative perception and will be instrumental in shaping the future of autonomous multi-robot systems.

\bibliographystyle{IEEEtran}
\bibliography{mybib}

\end{document}